\documentclass[11pt]{article}

\usepackage[expansion=false]{microtype}
\usepackage{graphicx}
\usepackage{booktabs}
\usepackage{amsmath}
\usepackage{amsfonts}

\usepackage{amssymb}
\usepackage{bm}
\usepackage{enumitem}
\usepackage{xspace}
\usepackage{makecell}
\usepackage{wrapfig}
\usepackage{multirow}
\usepackage{array}
\usepackage{float}
\usepackage{algorithm}
\usepackage{algpseudocode}
\usepackage{ziplab-tech-report}
\usepackage{hyperref}

\newcommand{\ours}{ZipTok3D}
\newcommand{\oursfull}{High-Fidelity 3D Tokenization with Compact Token Prefixes}

\definecolor{bestscoreblue}{RGB}{222,235,247}
\definecolor{secondscoregreen}{RGB}{228,242,222}
\definecolor{thirdscoreyellow}{RGB}{255,248,210}

\newcommand{\reporttitle}{\ours{}: \oursfull{}}
\newcommand{\reportshorttitle}{ZipTok3D}
\newcommand{\reportauthors}{%
  \mbox{Mingda Lin\,\textsuperscript{1,$\dagger$}},\quad
  \mbox{Weijie Wang\,\textsuperscript{1,$\dagger$*}},\quad
  \mbox{Zeyu Zhang\,\textsuperscript{1}},\quad
  \mbox{Bowen Cui\,\textsuperscript{1}},\quad
  \mbox{Yefei He\,\textsuperscript{1}}\\[-0.1em]
  \mbox{Haoyu Zhao\,\textsuperscript{1}},\quad
  \mbox{Yuanyu He\,\textsuperscript{1}},\quad
  \mbox{Donny Y. Chen\,\textsuperscript{2}},\quad
  \mbox{Feng Chen\,\textsuperscript{3,*}},\quad
  \mbox{Bohan Zhuang\,\textsuperscript{1}}%
}
\newcommand{\reportaffiliations}{%
  \textsuperscript{1}Zhejiang University \quad
  \textsuperscript{2}Monash University \quad
  \textsuperscript{3}University of Adelaide}
\newcommand{\reportdate}{September 2, 2026}
\newcommand{\reportconference}{}
\newcommand{\reportproject}{\href{https://forthloth.github.io/ziptok3d/}{https://forthloth.github.io/ziptok3d/}}
\newcommand{\reportemail}{%
  Weijie Wang: \href{mailto:wangweijie@zju.edu.cn}{wangweijie@zju.edu.cn};\quad
  Feng Chen: \href{mailto:chenfeng1271@gmail.com}{chenfeng1271@gmail.com}}
\newcommand{\reportorcid}{}
\newcommand{\reportkeywords}{3D tokenization, 3D representation learning, neural fields, iterative refinement}
\newcommand{\reportfootnote}{%
  \textsuperscript{$\dagger$} Equal contribution.\quad
  \textsuperscript{*} Corresponding authors.}
\newcommand{\reportpdfauthors}{Mingda Lin, Weijie Wang, Zeyu Zhang, Bowen Cui, Yefei He, Haoyu Zhao, Yuanyu He, Donny Y. Chen, Feng Chen, Bohan Zhuang}
\newcommand{\reportlogoheight}{0.62cm}

\techreportlabel{ZIP Lab Technical Report}
\techreportshorttitle{\reportshorttitle}
\techreportsetlogoheight{\reportlogoheight}
\techreportdate{\reportdate}
\techreportconference{\reportconference}
\techreportproject{\reportproject}
\techreportemail{\reportemail}
\techreportorcid{\reportorcid}
\techreportkeywords{\reportkeywords}
\techreportfootnote{\reportfootnote}

\techreportlogos{%
  \techreportlogo{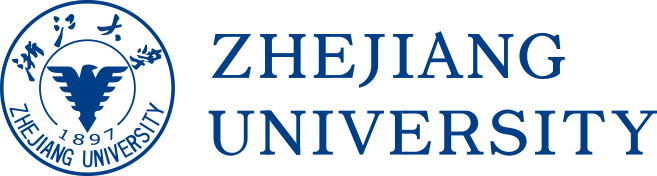}
  \techreportlogo{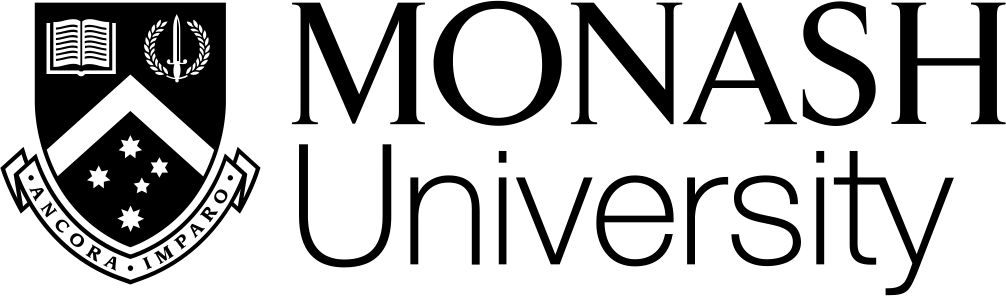}
  \techreportlogo{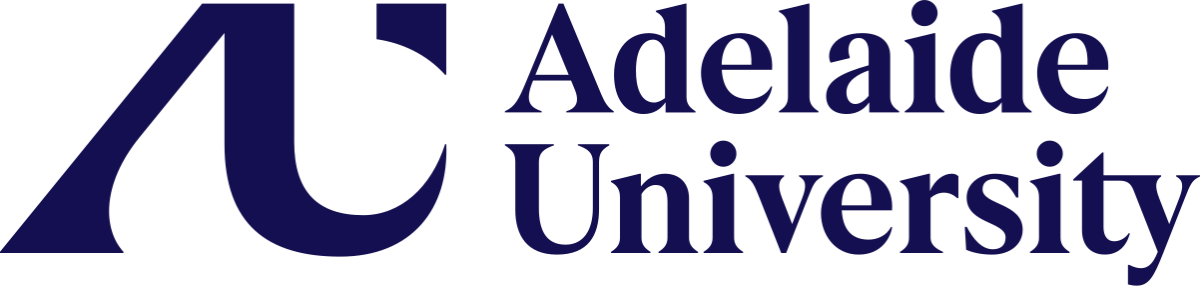}
}

\hypersetup{
  colorlinks=true,
  linkcolor=ziplabblue,
  citecolor=ziplabblue,
  urlcolor=ziplabblue,
  pdftitle={\reporttitle},
  pdfauthor={\reportpdfauthors},
  pdfsubject={ZIP Lab technical report},
  pdfkeywords={\reportkeywords}
}

\makeatletter
\renewenvironment{figure*}{\@float{figure}}{\end@float}
\renewenvironment{table*}{\@float{table}}{\end@float}
\makeatother

\AtBeginEnvironment{figure}{%
  \centering%
  \captionsetup{width=\textwidth}%
}
\AtBeginEnvironment{figure*}{%
  \centering%
  \captionsetup{width=\textwidth}%
}

\begin{document}

\ziplabmaketitle{\reporttitle}{\reportauthors}{\reportaffiliations}{}
\ziplabendtitle{Compact token sequences are essential for efficient 3D generation. However, existing 3D tokenizers typically organize latent representations either over spatial regions or as fixed-size sets of global tokens, both suffering sharp reconstruction degradation when compressed to extremely low token budgets. In this paper, we present \textbf{ZipTok3D}, a 3D tokenizer designed for high-fidelity reconstruction from extremely short token sequences. Its key idea is to organize object geometry into progressively informative global-token prefixes and unfold these compact representations through iterative decoding. Specifically, nested dropout randomly truncates the latent sequence after encoding during training and requires each retained prefix to reconstruct the complete object, thereby prioritizing essential geometric information in the leading tokens. The decoder then repeatedly applies a parameter-shared Transformer block to recover fine-grained geometry from each prefix without a separate generative sampling stage. 
With the same token dimension, ZipTok3D achieves
reconstruction quality comparable to the 32-token COD-VAE baseline using only
one token on ShapeNet and four on TRELLIS, yielding $32\times$ and $8\times$
shorter token sequences, respectively.
}
\thispagestyle{firstpage}

\section{Introduction}

Recent 3D generation methods rely on tokenizers to convert complex object
geometry into latent sequences, whose length substantially affects downstream
generative modeling cost~\citep{zhang20243d,deng2025efficient,dutt2026lost,li2026supervoxelgpt}.
 Existing 3D tokenizers either preserve spatial structure through locally anchored tokens~\citep{zhang20223dilg,deng2025efficient,li2026supervoxelgpt} or compress object-wide geometry into a fixed set of global tokens~\citep{zhang20233dshape2vecset,cho2025representing,zhang2025lagem}. While global representations produce shorter sequences, their reconstruction quality degrades sharply when the token budget is reduced to only a few tokens~\citep{zhang20233dshape2vecset,cho2025representing}. This exposes a fundamental tension between latent sequence length and geometric fidelity: \textit{can an entire 3D object be faithfully reconstructed from an extremely short token sequence?}


High-fidelity 3D reconstruction from only a few tokens poses a severe information bottleneck. With a moderate token budget, existing global 3D tokenizers can distribute complementary geometric information across multiple latent vectors. When the budget is reduced to only a few tokens, however, each token must summarize a substantially larger portion of the object, forcing global structure and fine-grained details to compete for severely limited latent capacity. Since these tokenizers optimize the complete latent set at a fixed budget~\citep{zhang20233dshape2vecset,cho2025representing}, they resolve this competition only implicitly through the final reconstruction objective, without explicitly determining which information should be preserved first. We therefore optimize a nested family of prefixes, requiring short prefixes to retain the object-wide geometry necessary for faithful reconstruction while allowing additional tokens to encode residual details.

\clearpage
\begin{wrapfigure}[14]{r}{.50\textwidth}
  \centering
  \includegraphics[width=\linewidth]{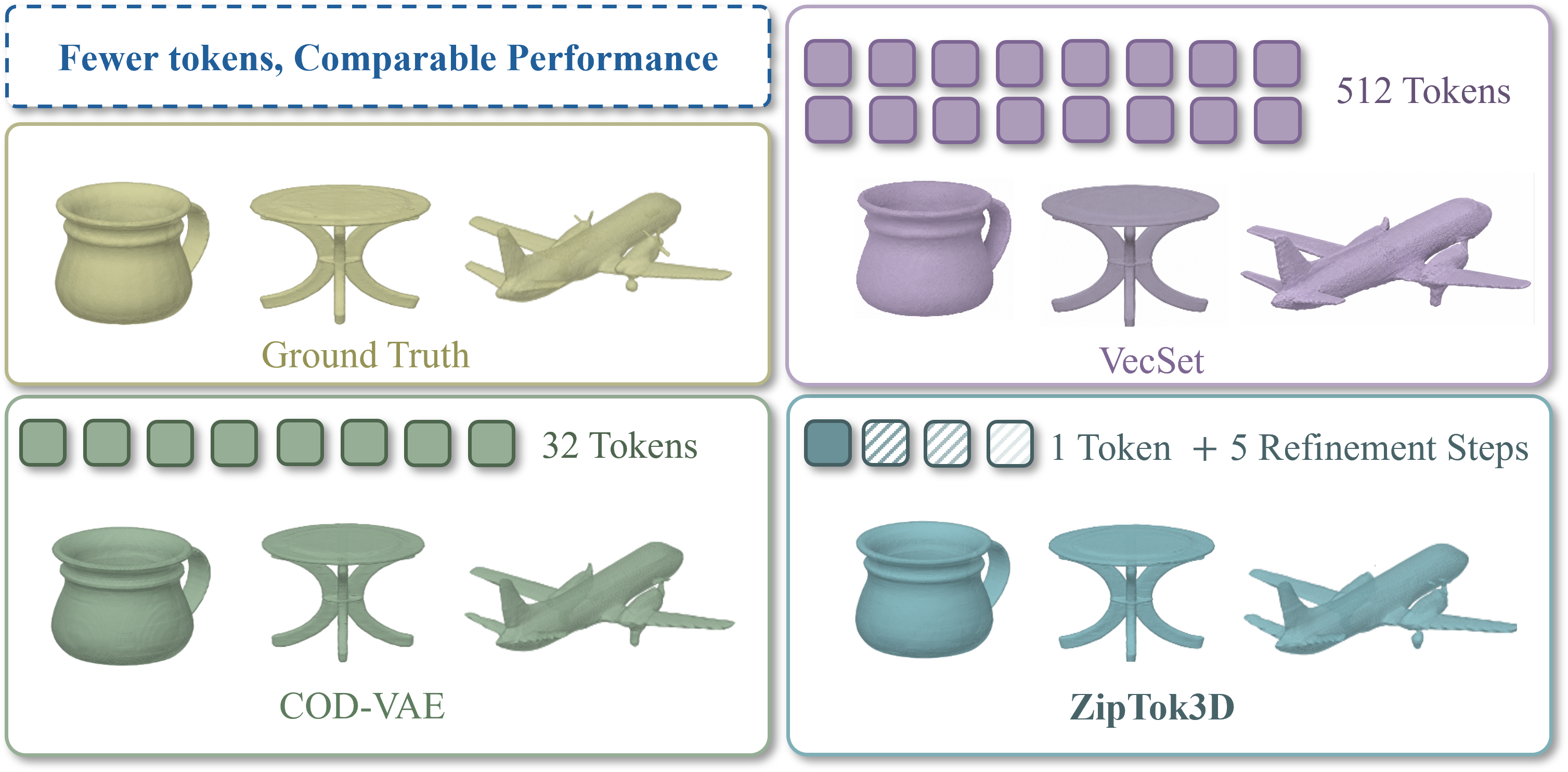}
  \captionsetup{width=\linewidth,font=small,aboveskip=2pt,belowskip=\baselineskip}
  \caption{Qualitative reconstruction at different token budgets.
  Ground-truth shapes are shown in the upper left. VecSet uses 512 global
  tokens, while COD-VAE uses 32. ZipTok3D reconstructs the same objects
  from the first token of its ordered latent sequence with five shared
  refinement steps.}
  \label{fig:teaser}
\end{wrapfigure}

Concentrating essential geometry into the leading tokens, however, addresses only the representation side of the problem. As the prefix becomes shorter, object-wide geometry is encoded in an increasingly compact form, placing a greater burden on the decoder to transform a few global vectors into a spatially detailed 3D representation. Conventional fixed-depth decoders perform this transformation through a single feed-forward process~\citep{zhang20233dshape2vecset,cho2025representing} and may not fully exploit the information contained in extremely short prefixes. Faithful low-token reconstruction therefore requires not only prioritizing what is encoded in the leading tokens, but also progressively unfolding their compact geometric information during decoding.

To address these challenges, we present \textbf{ZipTok3D}, a 3D tokenizer that jointly learns progressively informative token prefixes and iteratively decodes the geometry they contain. During training, nested dropout~\citep{rippel2014learning} exposes the tokenizer to varying prefix lengths of the encoded latent sequence and requires each sampled prefix to reconstruct the complete input, encouraging the leading tokens to preserve object-wide geometry while subsequent tokens provide additional details. To effectively unfold the compact information encoded in short prefixes, the decoder recurrently applies a parameter-shared Transformer block~\citep{dehghani2018universal,goyal2026elt} under intermediate reconstruction supervision, progressively refining the spatial representation without introducing step-specific parameters. Unlike methods that rely on generative completion~\citep{bachmann2025flextok}, ZipTok3D directly reconstructs the input geometry from each retained prefix without a separate sampling process. Using five refinement steps, ZipTok3D achieves reconstruction quality
comparable to 32-token COD-VAE from only one token on ShapeNet and four tokens
on TRELLIS, reducing representation length by $32\times$ and $8\times$,
respectively.

Our contributions are summarized as follows:
\begin{itemize}
    \item We introduce \textbf{ZipTok3D}, a 3D tokenizer designed for faithful reconstruction from extremely short token sequences.

    \item We couple nested prefix tokenization with parameter-shared iterative refinement, enabling the leading tokens to preserve object-wide geometry and progressively unfolding their compact information into detailed 3D representations.

    \item Experiments show that one ZipTok3D token approaches COD-VAE-32 on
    ShapeNet, while four tokens surpass it on all reported TRELLIS metrics,
    using $32\times$ and $8\times$ fewer tokens.

\end{itemize}


\section{Related Work}

\subsection{3D Latent Representations}

Spatially organized methods associate latent features with points, grids, or
hierarchical structures. LION uses hierarchical point-cloud features~\citep{zeng2022lion}, 3DILG adopts irregular grids~\citep{zhang20223dilg}, and OctFusion uses octrees~\citep{xiong2025octfusion}. XCube builds sparse voxel hierarchies~\citep{ren2024xcube}, while TRELLIS attaches features to occupied sparse-grid
cells~\citep{xiang2025structured}. Recent methods further compress such
representations through coarse voxel anchors, as in LATTICE~\citep{Lai_2026_CVPR}, or sparse geometry-and-appearance latents, as in O-Voxel~\citep{Xiang_2026_CVPR}. These representations preserve spatial support, but
reducing it generally requires coarser or sparser structures. Related feed-forward
3D representations likewise expose a compression--fidelity trade-off when forming
compact latent states~\citep{wang2026zpressor}.

Global methods instead compress object-wide geometry into compact token sets.
VecSet obtains neural-field latents through cross-attention~\citep{zhang20233dshape2vecset}, while COD-VAE progressively compresses point
features for triplane reconstruction~\citep{cho2025representing}. SceneTok
extends permutation-invariant token sets to multi-view scene modeling~\citep{asim2026scenetok}. Other approaches use token hierarchies~\citep{zhang2025lagem}, apply multiscale residual
quantization~\citep{zhang20243d}, or preserve surfaces via geometry-aware
sampling~\citep{chen2025dora}.

Compact codes also support diverse decoders and generative models. Shape Tokens
condition a flow-matching surface field~\citep{chang20243d}, Kyvo uses
quantized shape codes for multimodal autoregressive scene modeling~\citep{Sahoo_2026_CVPR}. FlashVDM accelerates VecSet-based generation
through diffusion distillation and efficient implicit decoding~\citep{Lai_2025_ICCV}, while Block3D studies block-wise diffusion for efficient
text-to-3D generation~\citep{cui2026block3d}. However, fixed-budget objectives do not explicitly concentrate reconstructive
information into short prefixes, so methods such as VecSet and COD-VAE can
degrade sharply at very small token budgets.

\subsection{Flexible-Length Tokenization}

Flexible-length tokenizers learn representations that remain decodable at
multiple sequence lengths. Nested dropout induces an information ordering by
randomly truncating latent sequences during training~\citep{rippel2014learning}. Related image tokenizers learn prefix-decodable or
causal one-dimensional representations~\citep{bachmann2025flextok,Wen_2025_ICCV,Chen_2026_CVPR}, while adaptive
methods allocate token budgets according to input complexity~\citep{duggal2025adaptive,yan2025elastictok,li2026adaptok,Xiong_2026_CVPR}.
VideoFlexTok learns coarse-to-fine video prefixes with a generative flow
decoder~\citep{atanov2026videoflextok}, and ReTok improves the utilization of
later tokens in nested-dropout representations~\citep{fu2026improving}.

Recent 3D methods vary sequence length through spatial or semantic
organization. OAT allocates octree tokens according to shape complexity~\citep{deng2025efficient}, while SuperVoxelGPT adapts supervoxel size to local
detail under a fixed generation order~\citep{li2026supervoxelgpt}. LoST orders
tokens by semantic salience, with early prefixes specifying overall shape and
later tokens adding instance details~\citep{dutt2026lost}. 
Its short prefixes
condition diffusion-based completion rather than recover the encoded instance
exactly. ZipTok3D instead learns nested reconstructive prefixes of global
latents and decodes them directly through shared iterative refinement, without
adaptive spatial partitioning or generative completion.

\subsection{Iterative Refinement}

A shared refinement module adds depth without step-specific parameters. The Universal Transformer
addresses the fixed depth of standard Transformers by recurrently applying
shared self-attention and feed-forward layers~\citep{dehghani2018universal}. ELT extends this idea to image and video
generation, using weight-shared Transformer loops and intra-loop
self-distillation to support different refinement depths~\citep{goyal2026elt}.

Related strategies have been used in 3D to recover geometric detail from
coarse or incomplete point clouds. The Cascaded Refinement Network recovers
details missing from coarse predictions through cascaded coarse-to-fine
refinement~\citep{wang2020cascaded}. RFNet reduces the parameter and memory
costs of dense completion by sharing operations across recurrent levels while
progressively increasing point density and preserving observed details~\citep{huang2021rfnet}. Together, these advances provide a useful basis for
decoding detailed geometry from highly compressed token sequences.

%


\clearpage
\section{Method}

\begin{figure}[H]
  \centering
  \includegraphics[width=\textwidth]{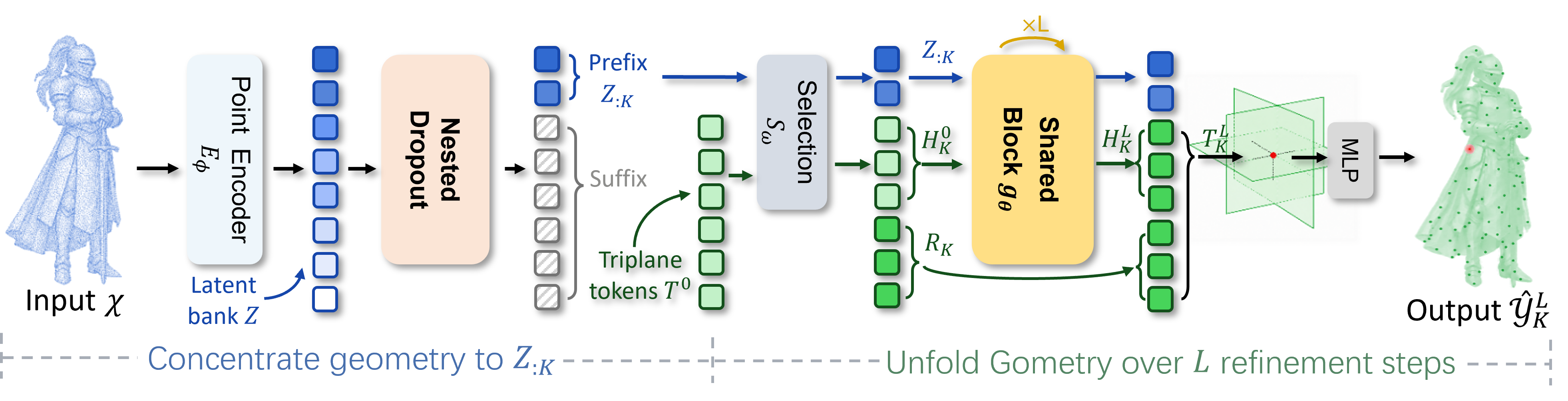}
  \caption{Overview of \ours{}. Nested dropout concentrates geometry into a
  reconstructive prefix $Z_{:K}$, while the parameter-shared block progressively
  unfolds it into triplane features over $L$ refinement steps. Blue and green
  denote latent and triplane tokens, respectively; hatched tokens indicate the
  masked suffix.}
  \label{fig:main}
\end{figure}

\subsection{Overview}

The central difficulty of low-token reconstruction is not compression alone.
It combines two questions that a fixed-budget autoencoder does not separate:
\emph{what geometry remains available after the latent sequence is shortened},
and \emph{how much computation is required to transform that compact
representation into a spatial field}. We expose these factors as two inference
controls. The retained prefix length $K$ sets the latent sequence length,
while the number of decoder iterations $L$ controls the computation used to
interpret that representation.

As shown in Figure~\ref{fig:main}, \ours{} changes how the latent sequence is
organized and replaces single-pass triplane decoding with a reusable update
rule. 
Let
$\mathcal{X}=\{x_i\}_{i=1}^{N}\subset\mathbb{R}^3$
denote the $N$ surface points sampled from an input shape. 
For an indexed collection of $N_q$ query points
$\mathcal{Q}=(q_1,\ldots,q_{N_q})\in\mathbb{R}^{N_q\times3}$, let
$\mathcal{Y}=(y_1,\ldots,y_{N_q})\in\{0,1\}^{N_q}$ denote the corresponding
ground-truth occupancy labels, where $y_j$ is the occupancy label of $q_j$.
A progressive point encoder $E_\phi$, parameterized by
$\phi$, maps $\mathcal{X}$ to a sequence of $M$ latent vectors~\citep{cho2025representing}:
\begin{equation}
Z=E_\phi(\mathcal{X})=(z_1,\ldots,z_M)
\in\mathbb{R}^{M\times d},
\qquad z_i\in\mathbb{R}^{d},
\label{eq:latent-bank}
\end{equation}
where $M$ is the maximum sequence length and $d$ is the shared latent width.
During training, nested dropout retains the prefix
$Z_{:K}=(z_1,\ldots,z_K)\in\mathbb{R}^{K\times d}$, where
$1\leq K\leq M$.

For decoding, the learnable triplane tokens $T^0$ first interact with the
retained prefix $Z_{:K}$ within the selection block
$\mathcal{S}_{\omega}$~\citep{cho2025representing}, which partitions the
triplane tokens into the selected state $H_K^0$ and the redundant tokens
$R_K$. This operation is denoted by
$(H_K^0,R_K)=\mathcal{S}_{\omega}(T^0,Z_{:K})$.
The retained prefix then conditions a parameter-shared Transformer block that
refines $H_K^0$ for $L$ iterations. After refinement, the updated state and
the redundant tokens are restored to a dense triplane representation. The
occupancy MLP queries this representation at the locations in $\mathcal{Q}$,
producing the final prediction $\hat{\mathcal{Y}}_K^L$.

\subsection{Learning a Reconstructive Prefix Code}

A reconstruction objective applied only to the complete latent sequence
constrains the information carried jointly by all tokens, but not how that
information is allocated across token positions. The decoder may therefore
rely on features distributed throughout the sequence, leaving its first few
tokens insufficient for reconstructing the complete object. We instead
optimize a nested family of prefixes across multiple token budgets. Each
shorter prefix is contained in every longer one, requiring the leading tokens
to support reconstruction at short lengths. Unlike arbitrary token dropout, this
nested structure defines a consistent representation at each supported length
and allows the sequence to be adjusted by truncation.

For each training example, the encoder first produces the complete latent
sequence $Z$ with maximum length $M=128$. We uniformly sample the retained
length from the exponentially spaced token budgets
\begin{equation}
\begin{gathered}
\mathcal{K}=\{1,2,4,8,16,32,64,128\},\\
K\sim\operatorname{Unif}(\mathcal{K}).
\end{gathered}
\label{eq:nested-prefix-sampling}
\end{equation}
During training, the suffix $Z_{K+1:M}$ is
masked from both triplane selection and iterative refinement, so the 
reconstruction objective must be realized using only $Z_{:K}$. 
At inference time,
the masked suffix is removed, allowing the same checkpoint to operate at every
trained token budget.


\subsection{Iterative Global-to-Spatial Refinement}

Prefix learning determines which evidence reaches the decoder, but it does not
make the inverse mapping from a few global vectors to a detailed 3D field easy.
With very small $K$, a single feed-forward pass must distribute object-level
information across many spatial locations and recover the occupancy boundary
in one transformation. Increasing the number of distinct decoder layers adds
both computation and parameters while fixing the amount of computation after training.

We instead decode each retained prefix through $L$ refinement steps. Starting
from the compact triplane state $H_K^0$ produced by the selection module, the
same parameter-shared Transformer block repeatedly updates the state while
conditioning on the fixed prefix:
\begin{equation}
H_K^\ell
=
g_\theta\!\left(H_K^{\ell-1},Z_{:K}\right),
\qquad
\ell=1,\ldots,L.
\label{eq:iterative-refinement}
\end{equation}
The parameters $\theta$ and the retained prefix $Z_{:K}$ remain unchanged
across iterations; only the triplane state is updated. Reusing the same block
therefore increases the effective decoding depth without introducing
step-specific parameters.

At refinement step $\ell$, the updated state and the redundant tokens retained
by the selection module can be restored to a complete triplane representation:
\begin{equation}
T_K^\ell
=
\mathcal{R}\!\left(H_K^\ell,R_K\right),
\label{eq:triplane-restoration}
\end{equation}
where $\mathcal{R}$ places the selected and redundant tokens back into their
corresponding triplane locations. Given the query points in $\mathcal{Q}$
defined above, the occupancy predictions at refinement step $\ell$ are
\begin{equation}
\begin{gathered}
\hat y_{K,j}^\ell
=
\sigma\!\left(
h_{\mathrm{occ}}(T_K^\ell,q_j)
\right),
\qquad j=1,\ldots,N_q,
\\
\hat{\mathcal{Y}}_K^\ell
=
\bigl(
\hat y_{K,1}^\ell,\ldots,\hat y_{K,N_q}^\ell
\bigr)
\in[0,1]^{N_q}.
\end{gathered}
\label{eq:iterative-occupancy}
\end{equation}
Here, $h_{\mathrm{occ}}$ is the shared occupancy MLP that predicts an
occupancy logit from the triplane features sampled at $q_j$, and $\sigma$
converts this logit into an occupancy probability.
The final reconstruction is $\hat{\mathcal{Y}}_K^L$, while
$\hat{\mathcal{Y}}_K^\ell$ for $\ell<L$ provides an intermediate prediction
along the same refinement trajectory. This recurrent process progressively
unfolds the geometric information concentrated in the retained prefix without
introducing an additional latent sampling stage.

\subsection{Training Objective}

Weight sharing alone does not make every refinement state a valid
reconstruction. If only the final iteration were supervised, earlier states
could remain unconstrained as reconstruction outputs. We therefore supervise
the final output together with one randomly sampled intermediate output from
the same refinement trajectory.

For a sampled prefix length $K$, the same retained prefix is used throughout
all $L$ refinement steps. 
We partition the query points in $\mathcal{Q}$ into two disjoint subsets:
the uniformly sampled volume queries $\mathcal{Q}_{\mathrm{vol}}$ and the
near-surface queries $\mathcal{Q}_{\mathrm{near}}$.
Let
$\mathcal{I}_{\mathrm{vol}}$ and $\mathcal{I}_{\mathrm{near}}$ denote their
corresponding index sets, with cardinalities $N_{\mathrm{vol}}$ and
$N_{\mathrm{near}}$, respectively, so that
$N_q=N_{\mathrm{vol}}+N_{\mathrm{near}}$. The reconstruction loss at
refinement step $\ell$ is
\begin{equation}
\mathcal{L}_{\mathrm{rec}}\bigl(\hat{\mathcal{Y}}_K^\ell,\mathcal{Y}\bigr)
=\frac{1}{N_{\mathrm{vol}}}\sum_{j\in\mathcal{I}_{\mathrm{vol}}}
\operatorname{BCE}\bigl(\hat y_{K,j}^\ell,y_j\bigr)
+\frac{\lambda_{\mathrm{near}}}{N_{\mathrm{near}}}
\sum_{j\in\mathcal{I}_{\mathrm{near}}}
\operatorname{BCE}\bigl(\hat y_{K,j}^\ell,y_j\bigr).
\label{eq:reconstruction-loss}
\end{equation}
Following COD-VAE \citep{cho2025representing}, we set
$\lambda_{\mathrm{near}}=0.1$.

To stabilize the joint optimization of triplane initialization and token
selection, we follow COD-VAE~\citep{cho2025representing} and retain its two
auxiliary losses:
\[
\mathcal{L}_{\mathrm{aux}}
=
\lambda_{\mathrm{init}}\mathcal{L}_{\mathrm{init}}
+
\lambda_{\mathrm{unc}}\mathcal{L}_{\mathrm{unc}}.
\]
We use
$\lambda_{\mathrm{init}}=0.5$ and $\lambda_{\mathrm{unc}}=0.001$.
Here, $\mathcal{L}_{\mathrm{init}}$ directly supervises the initial triplane
prediction, while $\mathcal{L}_{\mathrm{unc}}$ trains the uncertainty estimates
used for token selection.

To provide effective supervision for the intermediate states, we uniformly
sample a refinement step
$\ell\in\{1,\ldots,L-1\}$, where $L\geq2$. Its prediction
$\hat{\mathcal{Y}}_K^\ell$ receives direct reconstruction supervision and
intra-loop self-distillation~\citep{goyal2026elt}. Using the same query
weighting as the reconstruction loss, the distillation loss aligns the
intermediate and detached final occupancy predictions:
\begin{equation}
\mathcal{L}_{\mathrm{dist}}\bigl(\hat{\mathcal{Y}}_K^\ell,\mathrm{sg}(\hat{\mathcal{Y}}_K^L)\bigr)
=\frac{1}{N_{\mathrm{vol}}}\sum_{j\in\mathcal{I}_{\mathrm{vol}}}
\operatorname{BCE}\bigl(\hat y_{K,j}^\ell,\mathrm{sg}(\hat y_{K,j}^L)\bigr)
+\frac{\lambda_{\mathrm{near}}}{N_{\mathrm{near}}}
\sum_{j\in\mathcal{I}_{\mathrm{near}}}
\operatorname{BCE}\bigl(\hat y_{K,j}^\ell,\mathrm{sg}(\hat y_{K,j}^L)\bigr).
\label{eq:distillation-loss}
\end{equation}

Here, $\mathrm{sg}(\cdot)$ denotes stop gradient. The detached final
probability serves as a soft occupancy target for the intermediate prediction.

For the sampled token budget $K$ and intermediate depth $\ell$, the overall training objective is
\begin{equation}
\mathcal{L}_{K,\ell}=
\underbrace{\mathcal{L}_{\mathrm{rec}}\bigl(\hat{\mathcal{Y}}_K^L,\mathcal{Y}\bigr)+\mathcal{L}_{\mathrm{aux}}}_{\text{final and auxiliary supervision}}
+\beta\,\underbrace{\Bigl[\alpha_t\,\mathcal{L}_{\mathrm{rec}}\bigl(\hat{\mathcal{Y}}_K^\ell,\mathcal{Y}\bigr)+(1-\alpha_t)\,\mathcal{L}_{\mathrm{dist}}\bigl(\hat{\mathcal{Y}}_K^\ell,\mathrm{sg}(\hat{\mathcal{Y}}_K^L)\bigr)\Bigr]}_{\text{intermediate supervision}}.
\label{eq:training-objective}
\end{equation}
Here, $\beta=0.5$ controls the contribution of
intermediate supervision.
At optimizer step $t$, we set
$\alpha_t=\max(0,1-t/T_{\alpha})$, where $t$ counts completed optimizer
updates and $T_{\alpha}$ is the total number of scheduled updates after
accounting for distributed training and gradient accumulation.

\section{Experiments}

\subsection{Experimental Setup}

\subsubsection{Datasets}

We evaluate 3D reconstruction on ShapeNet-v2 (ShapeNet)~\citep{chang2015shapenet} and TRELLIS-500K (TRELLIS)~\citep{xiang2025structured}. For ShapeNet, we follow the preprocessing and split of COD-VAE~\citep{cho2025representing}; the evaluation split contains 1,283 objects from
55 categories.
For TRELLIS, we retain assets with polygonal meshes, apply the watertight
preprocessing of VecSet~\citep{zhang20233dshape2vecset}, and partition the data
with seed 42, yielding an evaluation split of 2,613 assets. Supplementary Sec.~S2 reports the source composition and exact asset
manifest for this split.

\subsubsection{Baselines}
For quantitative comparison, we consider 3DILG~\citep{zhang20223dilg}, VecSet~\citep{zhang20233dshape2vecset}, and COD-VAE~\citep{cho2025representing}, which support deterministic reconstruction under
fixed, shape-independent token budgets. All ShapeNet reconstruction results
follow the same test split and evaluation pipeline established by COD-VAE,
with additional operating points evaluated using the official implementations.
All TRELLIS results use the same split.

\subsubsection{Models and Training}

The tokenizer takes 2,048 surface points and produces at most 128 latent tokens
of width 512. At inference, \ours{} receives the actual
$K\times512$ prefix without suffix padding; COD-VAE uses the same token width.
The decoder reuses one six-layer Transformer block across refinement passes,
and the main results use five passes to produce a $128\times128$ triplane.
Supplementary Secs.~S3 provide detailed architecture specifications,
recurrent refinement inputs, and definitions of the inherited auxiliary
losses.

We train the tokenizer in FP16 with AdamW using a learning rate of $10^{-4}$,
weight decay $0.01$, and an effective batch size of 672 on four A800 GPUs.
ShapeNet and TRELLIS training run for 1,000 and 300 epochs, respectively. We
select checkpoints by the highest validation query IoU. Each tokenizer and
ablation configuration is trained once using seed 123456. Supplementary
Sec.~S2 specifies the optimizer parameters, per-GPU batch size, gradient
accumulation, learning-rate schedules, loss weights, and model-selection
protocol.

For class-conditioned generation, a causal second-stage VAE maps each
$K\times512$ tokenizer prefix to a $K\times32$ latent sequence without changing
its length. At the reported operating point, the EDM models the exact
$2\times32$ sequence without suffix padding, whereas COD-VAE-32 uses a
$32\times32$ stage-2 sequence. 
Supplementary provide the stage-2 training protocol and
detailed architectural specifications of the prefix VAE and EDM,
respectively.

\subsubsection{Evaluation}

Following COD-VAE, reconstruction is evaluated using query IoU over 500,000
volume queries and mesh CD and F1 computed from $128^3$ marching-cubes
reconstructions. For class-conditioned generation, we evaluate airplane, car,
chair, table, and rifle using 2,000 generated shapes per category and report
MMD-CD, COV-CD, and 1-NNA-CD. The diffusion models use 18 sampling steps,
whereas 3DILG uses its native 512-step autoregressive sampler.

Efficiency is measured using trained checkpoints on one NVIDIA H20 GPU with
batch size 16 and FP32 inference, with TF32 and Flash SDP disabled. \ours{}
always receives an exact prefix without suffix padding. Sampling throughput
covers latent generation, while full throughput additionally includes
tokenizer decoding and the complete $128^3$ occupancy-field query. 
Supplementary Secs.~S4 provide detailed evaluation settings,
reference-set construction, sampling parameters, query chunking, CUDA timing,
and peak-memory measurement.



Sampling throughput covers latent generation only, whereas full throughput
additionally includes decoding and the complete $128^3$ occupancy-field query
in chunks of 131,072 points. CUDA events are used for timing. Decoder latency
discards three warmup batches and averages ten measured batches; generation
efficiency discards five warmup batches and averages thirty measured batches.
Peak memory is the maximum allocated, rather than reserved, GPU memory during
measured full inference. Timing excludes model initialization, data loading,
CPU--GPU transfers, marching cubes, and file writing.

\clearpage

\noindent
\begin{minipage}{\textwidth}
\centering
\captionsetup{width=\linewidth,hypcap=false}
\captionof{table}{Reconstruction results on ShapeNet and TRELLIS under the common
evaluation protocol described in the experimental setup. Tok. denotes latent
sequence length, and IoU and F1 are percentages. All \ours{} operating points
for a dataset are evaluated from the same checkpoint using five refinement
steps. Best results within each dataset and metric are shown in bold, and
second-best results are underlined; ties at the displayed precision share the
same formatting. A dash denotes an unavailable result when the corresponding
public checkpoint was not released.}
\label{tab:reconstruction-main}
{\normalsize
\setlength{\tabcolsep}{5pt}
\renewcommand{\arraystretch}{1.05}
\begin{tabular}{@{}lccccccc@{}}
\toprule
& &
\multicolumn{3}{c}{ShapeNet}
& \multicolumn{3}{c}{TRELLIS} \\
\cmidrule(lr){3-5}\cmidrule(lr){6-8}
Method & Tok.
& IoU$\uparrow$ & CD$\downarrow$ & F1$\uparrow$
& IoU$\uparrow$ & CD$\downarrow$ & F1$\uparrow$ \\
\midrule

3DILG
& 512
& 95.9
& \underline{0.013}
& \textbf{98.0}
& -- & -- & -- \\

VecSet
& 32
& 87.8
& 0.021
& 91.3
& -- & -- & -- \\

VecSet
& 64
& 91.2
& 0.017
& 94.7
& -- & -- & -- \\

VecSet
& 512
& 96.3
& \underline{0.013}
& \textbf{98.0}
& 71.47
& 0.0249
& 88.81 \\

\addlinespace[2pt]

COD-VAE
& 2
& 77.7
& 0.032
& 80.9
& 45.85
& 0.0674
& 53.95 \\

COD-VAE
& 32
& \underline{97.1}
& \textbf{0.012}
& 97.8
& 75.25
& 0.0172
& 95.67 \\

COD-VAE
& 64
& \textbf{97.5}
& \textbf{0.012}
& \textbf{98.0}
& \textbf{75.75}
& 0.0168
& \textbf{95.98} \\

\midrule

\ours{}
& 1
& 96.8
& \textbf{0.012}
& 97.8
& 75.18
& 0.0168
& 95.81 \\

\ours{}
& 2
& 96.9
& \textbf{0.012}
& 97.8
& 75.22
& \underline{0.0167}
& 95.86 \\

\ours{}
& 4
& 96.9
& \textbf{0.012}
& \underline{97.9}
& \underline{75.31}
& \textbf{0.0166}
& \underline{95.92} \\

\bottomrule
\end{tabular}}

\end{minipage}

\subsection{Main Results}

\subsubsection{Reconstruction from Extremely Short Prefixes}

Table~\ref{tab:reconstruction-main} shows that \ours{} retains high-fidelity
geometry at sequence lengths where fixed-budget tokenizers deteriorate. On
ShapeNet, the one-token setting attains the same rounded CD and F1 values as
COD-VAE-32 and remains within 0.3 IoU points, while using a $32\times$ shorter
latent sequence. By contrast, COD-VAE degrades sharply at two tokens, showing
that the result does not follow from simply reducing the size of a fixed latent
set. On the more diverse TRELLIS split, the four-token setting yields a similar
query-IoU point estimate and improves CD and F1 over COD-VAE-32 with an
$8\times$ shorter latent sequence.

For this primary TRELLIS comparison, we use a paired, object-level
nonparametric percentile bootstrap with 20,000 resamples and seed 123456.
Relative to COD-VAE-32, \ours{}
with $K=4$ changes query IoU by $+0.06$ points
(95\% CI $[-0.08,+0.20]$), reduces mesh CD by $0.00058$
(95\% CI $[0.00039,0.00080]$), and increases mesh F1 by $0.25$ points
(95\% CI $[0.16,0.34]$). 
The query-IoU interval includes zero, whereas the CD and F1 intervals exclude
zero and favor \ours{}. Supplementary Sec.~S4 details the resampling
procedure and clarifies that these intervals reflect variation across
evaluation objects rather than independent training runs.

\begin{figure}[!t]
\centering
\captionsetup{skip=3pt}
\includegraphics[width=0.88\textwidth]{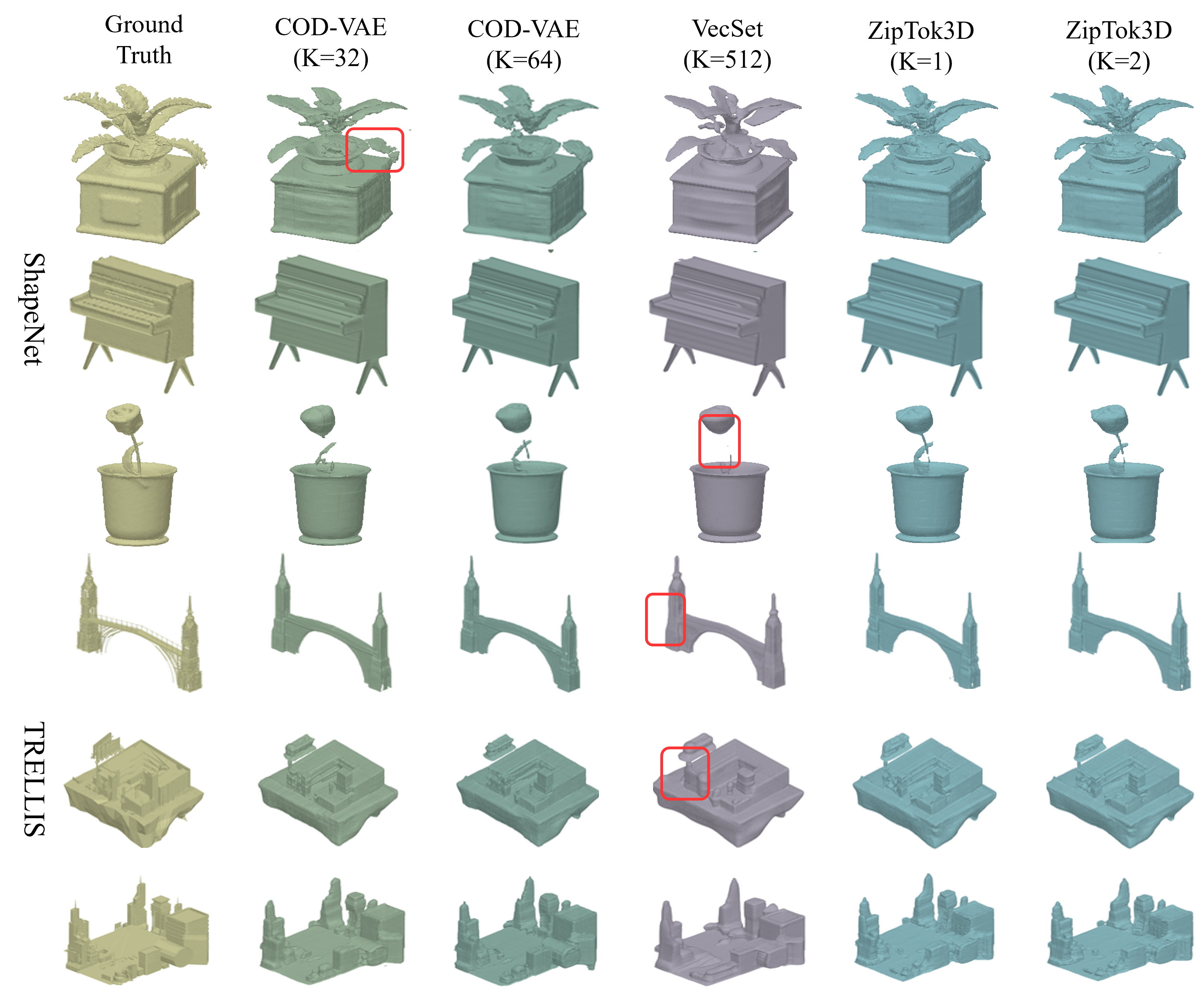}
\caption{Qualitative reconstructions on ShapeNet (top) and
\mbox{TRELLIS} (bottom). Columns show ground truth, COD-VAE with $K=32$ and
$K=64$, VecSet with $K=512$, and \ours{} with $K=1$ and $K=2$. Both \ours{}
settings use five refinement steps.}
\label{fig:qualitative}
\end{figure}

Figure~\ref{fig:qualitative} further shows that one- and two-token prefixes
preserve global part layouts and thin structures, including bench legs, flower
stems, and bridge towers, across both datasets.

\subsubsection{Class-Conditioned Generation}

We evaluate class-conditioned generation from an exact two-token
$2\times32$ latent code, with each generated code decoded using five shared
refinement passes.

\clearpage

\begin{table}[!t]
\centering
\caption{Class-conditioned generation on five ShapeNet categories.
All distribution metrics are CD-based. MMD is scaled by $10^{-3}$, while COV
and 1-NNA are percentages, with 50\% being ideal for 1-NNA. Samp. measures
latent generation alone, whereas Full additionally includes tokenizer decoding
and the complete $128^3$ occupancy-field query; both are reported in samples
per second. Mem. is peak allocated memory in GiB during the full pipeline.
Suffixes indicate stage-2 token count. Bold and underlined values denote the
best and second-best results, respectively.}
\label{tab:shapenet-generation}
{\normalsize\setlength{\tabcolsep}{2.5pt}
\begin{tabular}{@{}lcccccc@{}}
\toprule
& \multicolumn{3}{c}{Distribution}
& \multicolumn{3}{c}{Efficiency} \\
\cmidrule(lr){2-4}\cmidrule(l){5-7}
Method
& MMD$\downarrow$
& COV$\uparrow$
& 1-NNA$\downarrow$
& Samp.$\uparrow$
& Full$\uparrow$
& Mem.$\downarrow$ \\
\midrule

3DILG
& 6.270
& 58.16
& 61.49
& 0.24
& 0.22
& 26.18 \\

VecSet-32
& \underline{4.999}
& 84.51
& 54.32
& 28.39
& 6.59
& 25.68 \\

VecSet-64
& 5.090
& 84.89
& 55.04
& 25.60
& 6.11
& 26.19 \\

VecSet-512
& \textbf{4.807}
& \textbf{85.20}
& \underline{53.78}
& 4.27
& 2.14
& 33.24 \\

COD-VAE-32
& 5.020
& \underline{84.96}
& \textbf{53.11}
& \underline{46.02}
& \underline{36.14}
& \underline{2.26} \\

\midrule

\ours{}-2
& 5.142
& 84.91
& 54.21
& \textbf{50.62}
& \textbf{36.64}
& \textbf{2.20} \\

\bottomrule
\end{tabular}}

\end{table}











\FloatBarrier
Compared with COD-VAE-32, ZipTok3D-2 uses a $16\times$ shorter stage-2 latent
sequence while remaining close on all three distribution metrics. Under the
common efficiency protocol, it provides higher sampling throughput with
similar full throughput and peak memory, and is substantially faster than the
VecSet baselines across the full pipeline. These results show that the compact
prefix remains effective for downstream class-conditioned generation.

\clearpage

\begin{table}[!ht]
\centering
\caption{Ablation study on ShapeNet at $K=2$. \emph{w/o iter. ref.} denotes
the variant without iterative refinement, in which the six-layer block is
applied once. \emph{w/o inter. sup.} retains five refinement passes but
supervises only the final output. In the decoder column, $d\times p$ denotes
$d$ distinct Transformer layers applied for $p$ passes; shared layers are
counted once. Params and Lat. denote the number of trainable decoder parameters
in millions and decoder latency in milliseconds per shape, respectively. IoU
and F1 are percentages. Best results are shown in bold, and second-best results
are underlined; ties share the same formatting.}
\label{tab:mechanism-ablation}
{\normalsize\setlength{\tabcolsep}{2.5pt}
\begin{tabular}{@{}lccccccc@{}}
\toprule
& & & \multicolumn{3}{c}{Reconstruction}
& \multicolumn{2}{c}{Decoder Cost} \\
\cmidrule(lr){4-6}\cmidrule(l){7-8}
Variant & $K$ & Decoder
& IoU$\uparrow$ & CD$\downarrow$ & F1$\uparrow$
& Params$\downarrow$
& Lat.$\downarrow$ \\
\midrule

COD-VAE
& 2 & $12\times1$ & 77.7 & 0.032 & 80.9
& \underline{39.3} & \underline{1.47} \\

Prefix only
& 2 & $12\times1$ & 92.3 & 0.015 & 95.8
& \underline{39.3} & \underline{1.47} \\

w/o iter. ref.
& 2 & $6\times1$ & 91.9 & 0.017 & 94.2
& \textbf{23.5} & \textbf{0.98} \\

w/o inter. sup.
& 2 & $6\times5$ & \underline{96.6} & \underline{0.013}
& \underline{97.6}
& \textbf{23.5} & 2.65 \\

Full \ours{}
& 2 & $6\times5$ & \textbf{96.9} & \textbf{0.012}
& \textbf{97.8}
& \textbf{23.5} & 2.65 \\

\bottomrule
\end{tabular}}
\end{table}

\subsection{Ablation Study}
\subsubsection{Component Ablation}

We examine prefix training, shared iterative refinement, and intermediate
supervision through matched local controls on ShapeNet at $K=2$.
\emph{Prefix only} retains the original
12-layer, single-pass COD-VAE decoder and differs from COD-VAE only by applying
nested dropout during training. The remaining variants use the same six-layer
decoder as \ours{}. \emph{w/o iterative refinement} applies the block once, \emph{w/o inter.
sup.} applies it for five passes with supervision only on the final output, and
the full model additionally supervises an intermediate prediction.



At the matched $12\times1$ architecture and decoder cost, \emph{Prefix only}
improves all three reconstruction metrics over COD-VAE, supporting the effect
of nested prefix training in the original decoder. Within the six-layer
architecture, comparison between \emph{w/o iterative refinement} and
\emph{w/o inter. sup.} shows that repeated application of the shared block
provides the main additional improvement without adding decoder parameters,
although it increases latency. Intermediate supervision provides a modest
further improvement in final reconstruction without changing inference cost.
These results
identify prefix organization and shared refinement as the primary
contributors, with intermediate supervision providing a complementary benefit.


\subsubsection{Effect of Prefix Length and Refinement Depth}

\begin{figure}[!t]
    \centering
    \includegraphics[width=.82\textwidth]{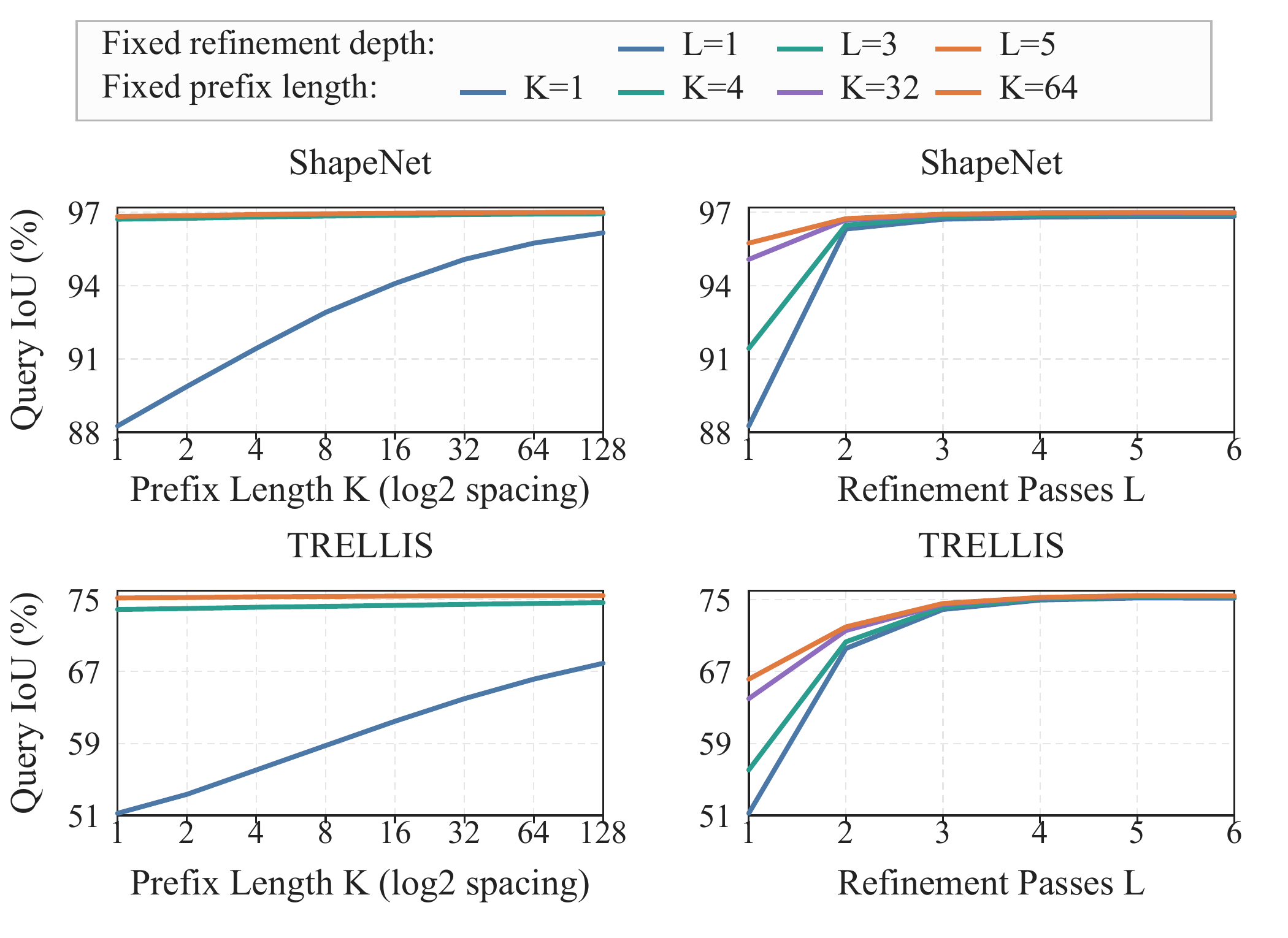}
    \caption{Effect of prefix length $K$ and refinement depth $L$ on query IoU
    for ShapeNet (top) and TRELLIS (bottom). The plots vary $K$ at fixed $L$
    (left) and $L$ at fixed $K$ (right).}
    \label{fig:budget-curves}
\end{figure}

We evaluate all supported combinations of prefix length and refinement depth
using one checkpoint per dataset. Figure~\ref{fig:budget-curves} shows the
query-IoU results at representative refinement depths.

\begin{figure}[!t]
    \centering
    \includegraphics[width=.82\textwidth]{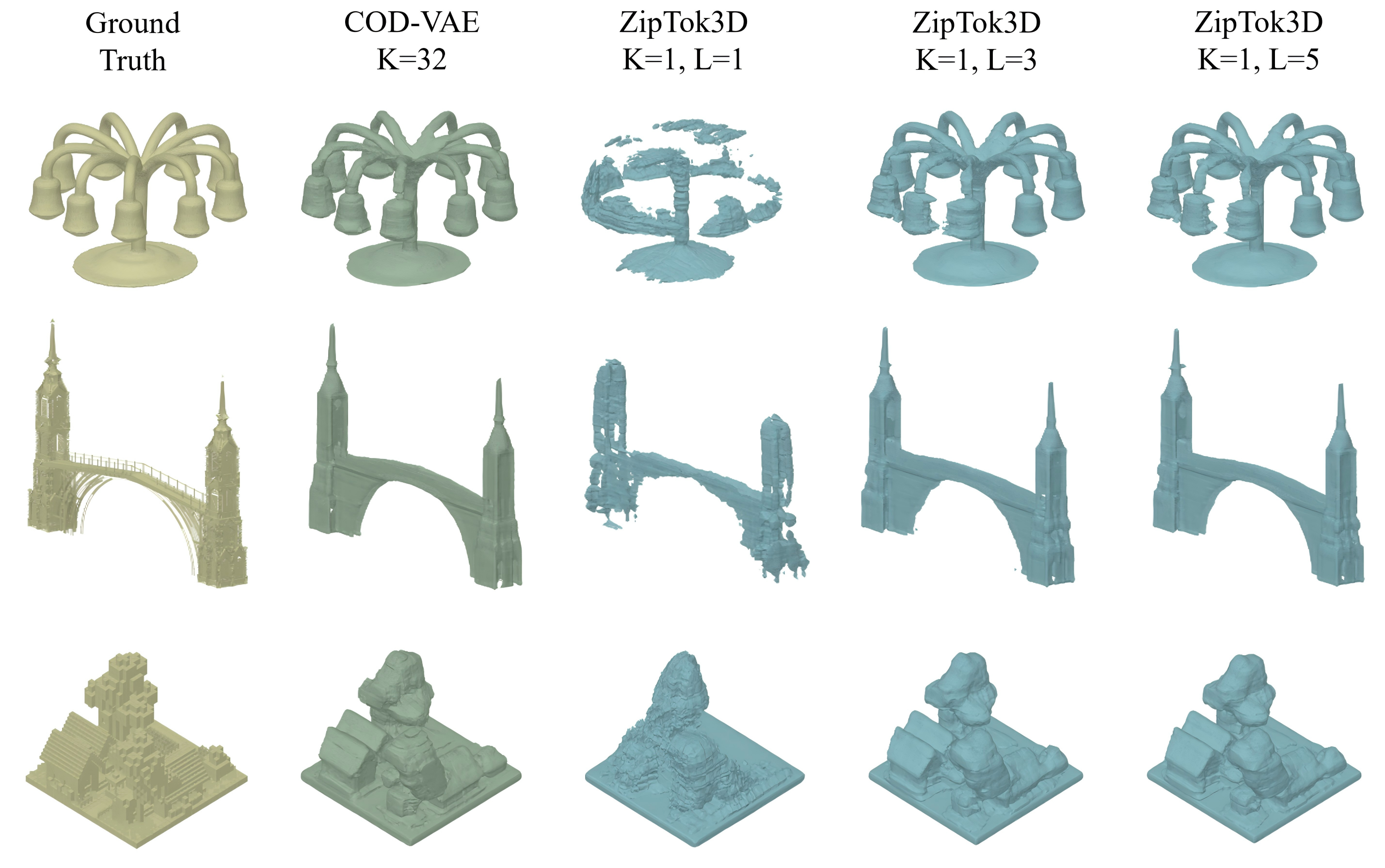}
    \caption{Qualitative refinement at $K=1$. Columns show ground truth,
COD-VAE with $K=32$, and ZipTok3D reconstructions with $K=1$ after
$L=1$, $L=3$, and $L=5$ refinement passes.}
    \label{fig:loop-effect}
\end{figure}

The two factors play complementary roles. Longer prefixes provide the largest
benefit under a single decoding pass, whereas performance becomes progressively
less sensitive to $K$ after three to five passes. Conversely, additional
refinement produces the largest gains at $K=1$ and $K=4$, while longer
prefixes already provide strong single-pass reconstructions. TRELLIS benefits
from deeper refinement than ShapeNet, but both datasets approach saturation by
$L=5$. Thus, additional tokens primarily compensate for shallow decoding,
whereas recurrent computation is most valuable in the extremely low-token
regime. The corresponding mesh CD and F1 curves exhibit the same interaction
pattern under the same settings (Supplementary Sec.~S1).

Figure~\ref{fig:loop-effect} illustrates this progression. With $L=1$, the
decoder captures portions of the global structure but may omit components or
produce fragmented surfaces. By $L=3$, it recovers the major structure and
produces a coherent reconstruction. The fifth pass primarily sharpens thin
structures, boundaries, and local details. This coarse-to-fine progression is
consistent with the diminishing quantitative gains near $L=5$.

These results suggest that the leading tokens retain the evidence needed to
recover global structure, but realizing it as complete geometry requires
repeated spatial refinement. Longer prefixes are therefore most beneficial
under shallow decoding, whereas deeper refinement more fully unfolds the
information concentrated in short prefixes.






\FloatBarrier
\section{Conclusion}
We presented \ours{}, a 3D tokenizer designed for high-fidelity reconstruction
from extremely short latent sequences. Nested prefix training encourages the
leading tokens to preserve object-wide geometry, while parameter-shared
iterative refinement progressively transforms this compact information into a
detailed spatial representation without introducing step-specific parameters.
Experiments show that one ZipTok3D token approaches the reconstruction quality
of 32-token COD-VAE on ShapeNet, while four tokens yield similar performance
on TRELLIS, reducing token counts by $32\times$ and $8\times$, respectively.
The ablation results confirm that prefix organization and iterative refinement
provide complementary benefits in the few-token regime. Our two-token stage-2
representation also supports class-conditioned generation with performance
close to COD-VAE-32, indicating that the compact prefixes remain useful for
downstream generative modeling. Future work will investigate generative models
that better exploit short latent sequences and adaptively allocate prefix
length and refinement depth according to each input.

\clearpage
\bibliographystyle{unsrtnat}
\bibliography{reference}

\clearpage
\appendix
\renewcommand{\thesection}{S\arabic{section}}
\renewcommand{\thesubsection}{S\arabic{section}.\arabic{subsection}}
\renewcommand{\thefigure}{\Alph{figure}}
\renewcommand{\thetable}{\Alph{table}}
\renewcommand{\theequation}{S\arabic{equation}}
\renewcommand{\theHequation}{app.\arabic{equation}}
\setcounter{section}{0}
\setcounter{figure}{0}
\setcounter{table}{0}
\setcounter{equation}{0}
\setcounter{secnumdepth}{2}

\section{Prefix-Length and Refinement-Depth Analysis}
\label{sec:full-sweeps}


\begin{figure}[H]
    \centering
    \captionsetup{skip=3pt}
    \begin{minipage}[t]{.485\textwidth}
        \centering
        \includegraphics[width=\linewidth]
        {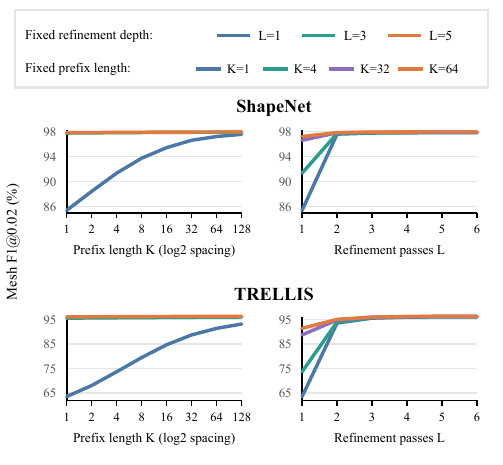}
        \par\smallskip
        \textbf{(a)} Mesh F1@0.02
    \end{minipage}
    \hfill
    \begin{minipage}[t]{.485\textwidth}
        \centering
        \includegraphics[width=\linewidth]
        {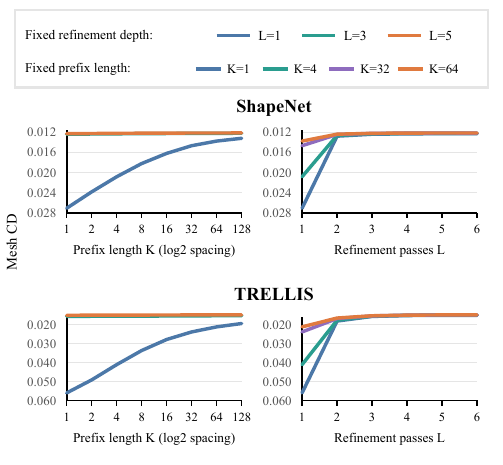}
        \par\smallskip
        \textbf{(b)} Mesh CD (lower is better)
    \end{minipage}
    \caption{Surface-metric interactions across prefix lengths $K$ and
    refinement depths $L$ on ShapeNet (top) and TRELLIS (bottom). The left
    panel reports mesh F1@0.02; the right panel reports mesh CD.}
    \label{fig:supp-surface-curves}
\end{figure}

\subsection{Complete Metric Sweeps}

To characterize how prefix length and refinement depth jointly affect
reconstruction, we evaluate all supported prefix lengths
$K\in\{1,2,4,8,16,32,64,128\}$ and refinement depths
$L\in\{1,2,3,4,5,6\}$. This extends the query-IoU analysis in the main paper
to the complete operating range with two surface metrics. All operating
points within each dataset use the same selected checkpoint, and the decoder
receives the exact $K\times512$ prefix without suffix padding.

Figure~\ref{fig:supp-surface-curves} provides complementary axis-wise views
of this interaction for mesh F1@0.02 and mesh CD. When refinement is shallow,
increasing $K$ yields the largest gains.
Conversely, when the prefix is short, increasing $L$ is most beneficial. This
asymmetric interaction is observed on both datasets and is consistent with the
query-IoU results in the main paper, showing that additional prefix tokens and
refinement passes contribute most strongly in different regimes.

\begin{figure*}[!t]
    \centering
    {\setlength{\tabcolsep}{1.5pt}
    \begin{tabular}{@{}cc@{}}
    \includegraphics[width=.40\textwidth]{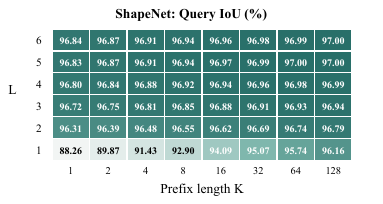} &
    \includegraphics[width=.40\textwidth]{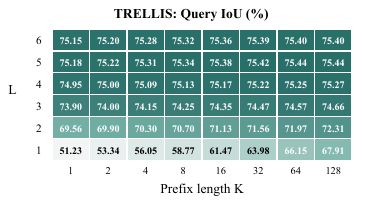} \\[-.70em]
    \includegraphics[width=.40\textwidth]{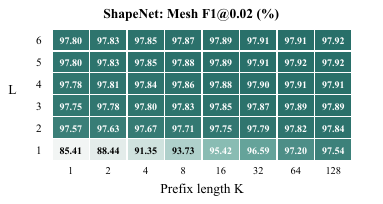} &
    \includegraphics[width=.40\textwidth]{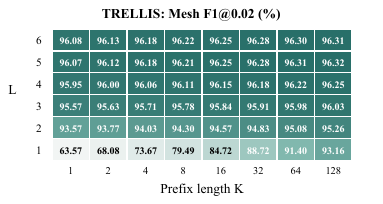} \\[-.70em]
    \includegraphics[width=.40\textwidth]{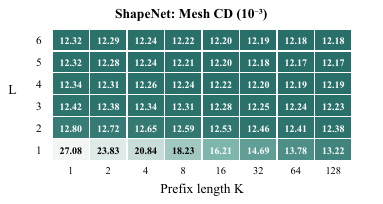} &
    \includegraphics[width=.40\textwidth]{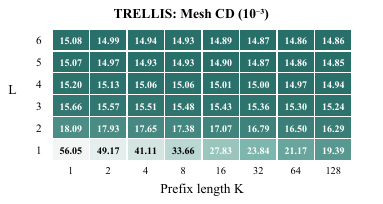}
    \end{tabular}}
    \par\vspace{-.35em}
    \includegraphics[width=.72\textwidth]{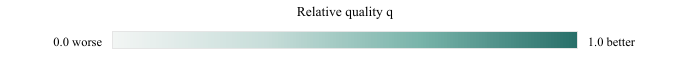}
    \par\vspace{-.25em}
    \caption{Complete ShapeNet (left) and TRELLIS (right) sweeps for
    refinement depths $L=1,\ldots,6$ and all supported prefix lengths. Rows
    report query IoU, mesh F1@0.02, and mesh CD. Cell values are dataset means;
    CD is reported in units of $10^{-3}$. To orient all metrics so that better
    values receive darker colors, we define the residual from the optimum as
    $r=100-m$ for percentage-valued IoU and F1 and as $r=m$ for CD. Within each
    dataset--metric panel, we apply standard min--max normalization to
    $-\log r$. If $r_{\min}$ and $r_{\max}$ are the minimum and maximum
    residuals over its displayed $(K,L)$ cells, the resulting color score is
    $q=\log(r_{\max}/r)/\log(r_{\max}/r_{\min})$. Thus, $q=1$ denotes the best
    cell and $q=0$ the worst. Darker colors indicate larger $q$.}
    
    \label{fig:supp-heatmaps}
\end{figure*}

Figure~\ref{fig:supp-heatmaps} reports the complete joint grids for query IoU,
mesh F1@0.02, and mesh CD. Across the evaluated range, the full sweeps confirm
that reconstruction is most sensitive to prefix length under shallow decoding,
whereas the benefit of additional refinement is concentrated at small $K$.
At the operating points used in the main paper, the differences between
$L=5$ and $L=6$ are small across all three metrics, supporting the use of five
refinement passes for evaluation.

\subsection{Per-Object Refinement Behavior}

Dataset-level mean gains do not show whether refinement benefits most objects
or only a small subset. Table~\ref{tab:supp-refinement-diagnostics} therefore
reports the fraction of objects whose IoU, CD, or F1 improves over an early
refinement transition, $L=1\!\rightarrow\!3$, and a later transition,
$L=3\!\rightarrow\!5$, at two short-prefix settings,
$K\in\{1,4\}$.
\par\medskip
\noindent
\begin{minipage}{\textwidth}
\centering
{\captionsetup{hypcap=false}\captionof{table}{Per-object improvement rates
across selected refinement intervals. An object is counted as improved if its
IoU or F1 strictly increases or its CD strictly decreases. Ties and regressions
remain in the denominator but are not counted as improvements. Values are
rounded to one decimal place.}}
\label{tab:supp-refinement-diagnostics}
{\normalsize
\setlength{\tabcolsep}{3pt}
\renewcommand{\arraystretch}{1.08}
\begin{tabular}{@{}llcccccc@{}}
\toprule
& & \multicolumn{3}{c}{ShapeNet} & \multicolumn{3}{c}{TRELLIS} \\
\cmidrule(lr){3-5}\cmidrule(lr){6-8}
$K$ & Transition
& IoU$\uparrow$ & CD$\downarrow$ & F1$\uparrow$
& IoU$\uparrow$ & CD$\downarrow$ & F1$\uparrow$ \\
\midrule
1 & $1\!\rightarrow\!3$ & 100.0 & 99.3 & 97.6 & 98.6 & 99.8 & 99.8 \\
1 & $3\!\rightarrow\!5$ & 88.2 & 81.1 & 58.1 & 91.2 & 91.8 & 84.7 \\
4 & $1\!\rightarrow\!3$ & 100.0 & 98.7 & 94.8 & 98.5 & 99.7 & 99.4 \\
4 & $3\!\rightarrow\!5$ & 89.5 & 81.5 & 55.7 & 90.5 & 90.7 & 83.4 \\
\bottomrule
\end{tabular}}
\end{minipage}
\par\medskip

Across both datasets, increasing $L$ from 1 to 3 improves a large majority of
objects on every metric, indicating that the aggregate gains are not driven by
a small subset. Increasing $L$ further from 3 to 5 still improves more than
half of the objects in every setting, but the improvement rates are lower and
more metric-dependent, with the weakest consistency for ShapeNet mesh F1.
Early refinement therefore provides broadly shared benefits, whereas later
refinement remains useful for a smaller and more metric-dependent fraction of
objects.

\FloatBarrier
\clearpage
\section{Experimental Setup Details}
\label{sec:experimental-details}

\subsection{Datasets, Preprocessing, and Splits}

\paragraph{ShapeNet.}
We use ShapeNetCore-v2 \citep{chang2015shapenet} with the 55-category
preprocessing and split definition released by 3DShape2VecSet
\citep{zhang20233dshape2vecset} and adopted by COD-VAE
\citep{cho2025representing}. The Stage-1 tokenizer, prefix VAE, and
class-conditional EDM share the same partition: 48,597 objects are used for
training, 2,592 for validation and model selection, and a disjoint 1,283-object
test set for final evaluation. Reconstruction is evaluated on the complete test
set, whereas generation metrics use its category-specific subsets for airplane,
car, chair, table, and rifle.
For each object, the released data provide separate pools of uniform volume
queries and near-surface queries, their occupancy labels, a normalization
scale, and a dense reference surface point cloud. Following COD-VAE, we
consolidate the records by split while preserving category and object
identifiers. All listed objects used in the reported experiments are
successfully converted. Query
coordinates are stored in FP16 for training and FP32 for both held-out splits,
while surface points remain in FP32.
During training, independent axis scales are sampled uniformly from
$[0.75,1.25]$, the shape is renormalized to the unit cube, and Gaussian noise
with standard deviation 0.005 is added to the surface points before clipping
to $[-1,1]$.  The same geometric scaling is applied to the occupancy queries.
Neither held-out split is augmented.

\paragraph{TRELLIS.}
TRELLIS-500K \citep{xiang2025structured} provides metadata for 500,777
training assets: 168,307 Objaverse-XL Sketchfab assets, 311,843 Objaverse-XL
GitHub assets, 4,485 ABO assets, 9,472 3D-FUTURE assets, and 6,670 HSSD assets.
Its separate 3,229-object Toys4K evaluation set is not used.  We merge
supported polygonal components and apply the
point-cloud-utils watertight conversion used by the VecSet preprocessing
pipeline at resolution 50,000 with seed 0.  The watertight mesh is centered at
its bounding-box center and scaled so that its maximum vertex radius is one.
Each successful object is represented by 100,000 surface points, 500,000
volume queries sampled uniformly from the radius-$\sqrt{3}$ sphere, and
500,000 near-surface queries divided equally between Gaussian perturbation
scales 0.005 and 0.05. Occupancy is one where the signed distance is
nonpositive and zero otherwise.
Per-object sampling is deterministic from seed 42 and the object identifier.
Inputs above 50,000 faces are decimated to that target before watertight
conversion when decimation succeeds. The preprocessing additionally sets a
900-second timeout and a 32-GiB memory limit per worker. An asset is excluded
if it lacks a supported,
nonempty polygonal mesh, contains nonfinite geometry, or fails mesh conversion,
normalization, surface sampling, signed-distance evaluation, or the stated
resource limits.

With seed 42, the successfully preprocessed assets are partitioned into 97\%
training, 2\% validation, and 1\% test subsets. The resulting test split
contains 2,613 objects: 57 ABO, 856 Objaverse-XL GitHub, and 1,700
Objaverse-XL Sketchfab assets. The anonymous code release provides the source
category and exact object identifier of every asset in this test split, which
is used for all reconstruction, bootstrap, and adaptive-budget analyses.

\subsection{Baseline Sources and Operating Points}

We compare with 3DILG \citep{zhang20223dilg}, VecSet
\citep{zhang20233dshape2vecset}, and COD-VAE
\citep{cho2025representing}.  These methods support deterministic
reconstruction at fixed, shape-independent token budgets.  All ShapeNet
comparisons use the common 1,283-object test split and the COD-VAE
metric implementation. The ShapeNet COD-VAE-32/64 checkpoints are inherited
from the official COD-VAE release, whereas COD-VAE-2 is trained using the
released implementation. On TRELLIS, COD-VAE-2/32/64 are initialized from
their corresponding ShapeNet checkpoints and fine-tuned on the common
preprocessed training pool. These checkpoints are selected by the highest mean
query IoU on the same validation split used by ZipTok3D.
VecSet-512 uses the inherited ShapeNet model without TRELLIS-specific training.
TRELLIS results are not reported for 3DILG or the shorter VecSet operating
points. All reported TRELLIS methods use the common test split and
evaluation protocol described above.

The reconstruction comparisons use 512 tokens for 3DILG; 32, 64, and 512
tokens for VecSet; and 2, 32, and 64 tokens for COD-VAE.  ZipTok3D is evaluated
with exact prefixes of 1, 2, and 4 tokens.  The class-conditioned generation
comparison uses the native stage-2 latent lengths of the baselines and the
exact two-token stage-2 representation of ZipTok3D.

\subsection{Stage-1 Optimization and Model Selection}
\label{sec:optimization}

For both datasets, each training example samples 2,048 points from the stored
surface pool as encoder input and separately samples 4,096 volume queries and
4,096 near-surface queries for reconstruction supervision. We optimize the
Stage-1 tokenizer with AdamW using $\beta_1=0.9$, $\beta_2=0.999$,
$\epsilon=10^{-8}$, weight decay 0.01, and an initial learning rate of
$10^{-4}$. Training uses FP16 mixed precision and clips the global gradient
norm at 0.5. Each of four A800 GPUs processes 56 examples, and gradients are
accumulated over three steps, giving an effective batch size of 672.

ShapeNet training runs for 1,000 epochs with a 50-epoch linear warmup followed
by cosine learning-rate decay. TRELLIS training runs for 300 epochs at a
constant learning rate. Every configuration is trained once with seed 123456.
For ShapeNet, checkpoints are selected by the highest mean query IoU on the
2,592-object validation split. For TRELLIS, the same criterion is evaluated on
the validation split. The implementation
uses Python 3.9, PyTorch 2.1.0, and CUDA 12.2. Training-time refinement and
supervision are specified in Section~\ref{sec:training-supervision}.

\subsection{Prefix VAE Optimization and Model Selection}

The prefix VAE uses AdamW with learning rate $10^{-4}$,
weight decay 0.01, global batch size 376, FP16 mixed precision, and gradient
clipping at 0.5.  It is trained for 100 epochs, with the learning rate halved at
epochs 60, 70, 80, and 90. It is trained on the same 48,597-object ShapeNet
training split as the Stage-1 tokenizer. Training uses four A800 GPUs and seed
123456. We select the checkpoint with the highest query IoU on the shared
2,592-object ShapeNet validation split.

\subsection{EDM Optimization and Model Selection}

The EDM uses AdamW with learning rate $10^{-4}$, weight decay 0.05, global
batch size 256, FP16 mixed precision, and gradient clipping at 1.0. Its
1,000-epoch schedule uses a 40-epoch linear warmup followed by cosine decay to
$10^{-6}$. Training uses four A800 GPUs and seed 123456. We set
$p_{\mathrm{mean}}=-1.2$, $p_{\mathrm{std}}=1.2$, and
$\sigma_{\mathrm{data}}=1$. The EDM is trained on latent arrays cached from the
shared ShapeNet training split, and its checkpoint is selected by the lowest
denoising loss on the shared validation split.

\section{Architecture and Objective Specifications}
\label{sec:architecture}

\subsection{Stage-1 Tokenizer Architecture}

Unless noted otherwise, the Stage-1 components follow the released COD-VAE design
\citep{cho2025representing}. ZipTok3D introduces nested prefix training, the
shared refinement block, and intermediate supervision.

The point encoder receives 2,048 surface points and forms 512 patches by
farthest-point sampling. It has four progressive blocks, each containing three
self-attention layers. The encoder width is 512, with eight attention heads,
a GEGLU MLP ratio of 4, and dropout 0.1. It produces a latent bank with maximum
length $M=128$ and width 512. The initial latent queries are the point
embeddings of 128 farthest-point-sampled surface positions, without a separate
learned latent-position table.

For each training example, one prefix length is sampled uniformly from
$\{1,2,4,8,16,32,64,128\}$. All suffix positions are masked from every decoder
attention operation. The decoder starts from 768 learned triplane tokens
$T^0$, corresponding to three 32-channel planes at spatial resolution
$128\times128$ with $8\times8$ patches. To predict occupancy, features are
bilinearly sampled from the three planes, summed, and processed by a
$32\!\rightarrow\!32\!\rightarrow\!1$ MLP with GELU.

\subsection{Triplane Initialization and Selection}

The selection block $\mathcal{S}_{\omega}$ first applies one cross-attention
layer between the 768 learned triplane tokens $T^0$ and the retained prefix
$Z_{:K}$. A LayerNorm--MLP uncertainty head assigns one score to each triplane
token. Hard descending top-$k$ selection retains the highest-scoring 25\%, or
192 tokens, as $H_K^0$. The remaining 576 tokens form the bypassed state
$R_K$ and retain their spatial indices for subsequent triplane restoration.
Selection is performed once before recurrent refinement rather than repeated
at every pass.

The ranking indices are discrete, so gradients do not propagate through the
top-$k$ ordering itself. Reconstruction gradients propagate through the
selected token features, while the uncertainty head receives direct
supervision from $\mathcal{L}_{\mathrm{unc}}$ defined below. The selected
indices are retained to restore the refined tokens to their original triplane
locations.

\subsection{Shared Refinement and Restoration}

The shared refinement module is a six-layer Transformer block of width 512
with eight attention heads, GEGLU MLP ratio 2, and dropout 0.1. It has no
causal mask, pass embedding, or pass-specific parameters. The recurrent
sequence at pass $\ell$ is
\begin{equation}
U_K^{\ell}=
\bigl[H_K^{\ell-1};Z_{:K}\bigr]
\in\mathbb{R}^{(192+K)\times512},
\end{equation}
where $H_K^{\ell-1}$ is the selected state and $Z_{:K}$ is the retained prefix.
The shared block processes this sequence, but only its first 192 outputs define
$H_K^{\ell}$. The prefix is reinserted unchanged at every pass, while the
bypassed tokens $R_K$ remain fixed outside the shared block.

After the final pass, the updated selected tokens are scattered to their
recorded spatial locations and combined with the 576 bypassed tokens at their
original locations. A linear residual head maps each restored spatial token to an
$8\times8\times32$ patch. The resulting residual is modulated by the predicted
uncertainty and added to the corresponding initial triplane patch.

\subsection{Auxiliary Losses and Intermediate Supervision}
\label{sec:training-supervision}

The two inherited COD-VAE auxiliary losses stabilize triplane initialization
and uncertainty-based selection. Let $a_j^0$ be the initial occupancy logit for
query $j$, $\hat y_j^0=\sigma(a_j^0)$ its probability, and
$\hat{\mathcal{Y}}^0=(\hat y_j^0)_{j=1}^{N_q}$ the corresponding probability
vector. The initialization loss applies the reconstruction criterion to this
initial prediction:
\begin{equation}
\mathcal{L}_{\mathrm{init}}
=\mathcal{L}_{\mathrm{rec}}(\hat{\mathcal{Y}}^0,\mathcal{Y}).
\end{equation}
The uncertainty target is the detached pointwise binary cross-entropy of the
initial prediction.  Following COD-VAE, it is clipped and normalized to
$[0,1]$ before regression:
\begin{align}
r_j & =
\frac{\operatorname{clip}(\operatorname{BCE}(\hat y_j^0,y_j)-0.01,0,0.99)}{0.99},\\
\mathcal{L}_{\mathrm{unc}}
&=\frac{1}{N_q}\sum_{j=1}^{N_q}(u_j-\operatorname{sg}(r_j))^2,
\end{align}
where $u_j$ is the uncertainty field decoded at the query. We use
$\lambda_{\mathrm{init}}=0.5$ and $\lambda_{\mathrm{unc}}=0.001$.

Stage-1 training unrolls six refinement passes. One intermediate depth is
sampled uniformly from $\{1,\ldots,5\}$ for each minibatch, while prefix
lengths are sampled independently for individual examples. The intermediate
term uses $\beta=0.5$, and its reconstruction and distillation components use
the same volume and near-surface weights, 1.0 and 0.1, as the final
reconstruction loss. Intermediate logits and detached final logits are
clamped to $[-30,30]$ before computing the distillation loss. The schedule
$\alpha_t$ is updated once per optimizer step and remains fixed across the
three microbatches accumulated into that update. Its endpoint $T_\alpha$ is
the scheduled optimizer-step count after distributed sharding, dropping
incomplete minibatches, and gradient accumulation. No additional
triplane-feature distillation term is used.

\subsection{Stage-2 Prefix VAE}

The prefix VAE accepts up to 16 ordered stage-1 tokens.  A LayerNorm and linear
projection independently map each 512-dimensional token to the mean and log
variance of a 32-dimensional diagonal Gaussian posterior.  The log variance is
clamped to $[-30,20]$, and training samples from the posterior by the
reparameterization trick.  A linear projection maps the sampled sequence back
to width 512, after which learned positional embeddings and a 12-layer
Transformer reconstruct the stage-1 token features.  This Transformer uses
eight heads, GEGLU MLP ratio 4, and dropout 0.1.  Its causal mask is
left-to-right:
\begin{equation}
A_{ij}=\begin{cases}
0, & j\leq i,\\
-\infty, & j>i,
\end{cases}
\end{equation}
so the reconstruction at position $i$ cannot use a later token.

Let $\widetilde Z$ denote a posterior sample and
$\bar Z=D_{\psi}(\widetilde Z)$ the reconstructed stage-1 token sequence.  The
stage-1 tokenizer is frozen while the prefix VAE is jointly optimized for
$\mathcal{B}=\{1,2,4,8,16\}$ with weights
$w_K\in\{1.2,1.1,1.0,0.9,0.8\}$.  Its objective is
\begin{equation}
\begin{aligned}
\mathcal{L}_{\mathrm{pVAE}}
&=\sum_{K\in\mathcal{B}}
\frac{w_K}{\sum_{K'\in\mathcal{B}}w_{K'}}
\Bigl[\operatorname{MSE}\!\left(\bar Z_{:K},\operatorname{LN}(Z_{:K})\right)
+\mathcal{L}_{\mathrm{rec}}\!\left(F_6(\bar Z_{:K}),\mathcal{Y}\right)\Bigr]
\\[-0.2em]
&\quad+10^{-3}D_{\mathrm{KL}}\!\left(
q_{\psi}(\widetilde Z\mid Z_{:16})\,\|\,\mathcal{N}(0,I)
\right).
\end{aligned}
\end{equation}
where $F_6$ denotes the frozen stage-1 decoder and occupancy head evaluated
with six refinement passes.  The feature and occupancy terms both have unit
coefficients; the volume and near-surface weighting inside
$\mathcal{L}_{\mathrm{rec}}$ is given in
Section~\ref{sec:training-supervision}.
Causality ensures that every term for budget $K$ depends only on the
corresponding stage-2 prefix. The six-pass decoder is used only to supervise
prefix VAE training. During generation evaluation, sampled codes are decoded
using five refinement passes.

\subsection{Stage-2 Class-Conditional EDM}

We cache deterministic posterior means for the full $16\times32$ latent array,
normalize them using training-set channel statistics, and train the
class-conditional EDM on arrays from all 55 ShapeNet training categories.
Left-to-right causal self-attention ensures that the denoising prediction at
each position depends only on the corresponding noisy prefix. At the reported
two-token operating point, sampling therefore instantiates and denoises only
the first two latent positions. The EDM and prefix VAE decoder thus operate on
an exact $2\times32$ tensor without suffix padding.
The EDM follows the latent-array design used by
VecSet and COD-VAE \citep{zhang20233dshape2vecset,cho2025representing}: it has
width 384, 16 Transformer blocks, eight 48-dimensional heads, GEGLU MLP ratio
4, no dropout, learned token-position embeddings, and a left-to-right causal
self-attention mask. A
256-dimensional sinusoidal noise embedding is projected to width 384 and
provides scale and shift to the adaptive LayerNorms in every block.  The class
index spans all 55 ShapeNet categories. It is mapped to a learned
384-dimensional embedding and supplied as a one-token key and value context
through cross-attention in every block; token self-attention therefore models
the latent sequence while class conditioning is injected separately.

For a normalized latent array $x$, category $c$, and
$\log\sigma\sim\mathcal{N}(p_{\mathrm{mean}},p_{\mathrm{std}}^2)$, the EDM
objective is
\begin{equation}
\begin{aligned}
\mathcal{L}_{\mathrm{EDM}}
&=\mathbb{E}_{x,c,\sigma,\epsilon}\!\left[
\frac{\sigma^2+\sigma_{\mathrm{data}}^2}
{(\sigma\sigma_{\mathrm{data}})^2}
\,\left\|D_{\theta}(x+\epsilon;\sigma,c)-x\right\|_2^2
\right],
\\[-0.2em]
&\qquad\epsilon\sim\mathcal{N}(0,\sigma^2I).
\end{aligned}
\end{equation}
All latent positions present at an operating point are denoised in parallel;
there is no token-autoregressive sampling loop.

\section{Evaluation Protocols}
\label{sec:evaluation}

\subsection{Reconstruction Metric Settings}

For both datasets, the encoder receives 2,048 points sampled from the stored
surface pool of the target object. The task is therefore deterministic
reconstruction from sampled target surfaces rather than completion from
partial observations. For query IoU, the resulting representation is decoded
at all 500,000 stored volume-query locations per object.

For mesh evaluation, the decoder is queried on a dense $128^3$ occupancy grid.
We extract the zero-logit level set, equivalently the 0.5-probability
isosurface, with marching cubes. We then sample 100,000 points from the
reconstructed mesh and compare them with the complete stored reference surface
point cloud. Chamfer distance is the sum of the two mean bidirectional
Euclidean nearest-neighbor distances. At threshold 0.02, precision and recall
are the fractions of reconstructed and reference points, respectively, lying
within the threshold of the other surface. F1 is their harmonic mean.

\subsection{Class-Conditioned Generation Metric Settings}

Following COD-VAE \citep{cho2025representing}, we evaluate airplane, car,
chair, table, and rifle. We generate 2,000 shapes for each category and report
minimum matching distance (MMD-CD), coverage (COV-CD), and 1-nearest-neighbor
accuracy (1-NNA-CD). For a given category, the reference set $S_r$ contains all
objects of that category in the 1,283-object ShapeNet test split. From the 2,000
generated candidates, a deterministic subset of $5|S_r|$ shapes is used for
MMD-CD and COV-CD, while a subset of $|S_r|$ shapes is used for 1-NNA-CD. The
same reference sets and generated-set construction are used for every method.

Each mesh is represented by 2,048 points sampled from its surface. Using CD as
the set distance, MMD averages the distance from each reference shape to its
nearest generated shape. COV is the fraction of reference shapes selected as a
nearest neighbor of at least one generated shape. The fixed 5:1
generated-to-reference ratio makes MMD-CD and COV-CD comparable across methods.
1-NNA is the leave-one-out nearest-neighbor classification accuracy on the
union of equally sized reference and generated sets, avoiding sample-count
imbalance between the two sets. We report the unweighted mean of each metric
over the five categories.

VecSet and COD-VAE use 18 diffusion steps, whereas 3DILG uses its native
512-step autoregressive sampler.  Our EDM uses 18 Heun steps with $\rho=7$,
$\sigma_{\max}=80$, and $\sigma_{\min}=0.002$, without condition dropout or
classifier-free guidance.

\subsection{Efficiency Measurement Protocol}

All efficiency measurements use the corresponding trained checkpoints on one
NVIDIA H20 GPU with batch size 16 and FP32 inference.  TF32 and Flash SDP are
disabled.  CUDA events measure GPU execution after warmup; model loading,
data loading, CPU--GPU transfers, marching cubes, and file output are excluded.
For reconstruction, full inference begins with the input point cloud and
includes point encoding, tokenizer decoding, and all $128^3$ occupancy queries
in chunks of 131,072 points. Decoder latency instead begins with an already
available stage-1 prefix and ends when the complete triplane has been
produced; it excludes point encoding and occupancy queries.

For generation, sampling throughput measures EDM latent generation alone.
Full throughput begins from the same latent noise and additionally includes
prefix VAE decoding, five-pass stage-1 tokenizer decoding, and the complete
$128^3$ occupancy-field query. Thus the point encoder is part of reconstruction
full inference but not of the generation pipeline. We discard three warmup
batches and average ten measured batches for reconstruction and decoder
latency; generation uses five warmup and thirty measured batches. For a
measured mean batch latency $t$ in milliseconds,
\begin{equation}
\mathrm{throughput}=\frac{16\times1000}{t}.
\end{equation}
Peak allocated CUDA memory is measured after resetting the peak-memory counter
following warmup. It is measured over the corresponding full pipeline;
reserved memory is not used.

\subsection{Reconstruction Efficiency Results}

\begin{table}[!t]
\centering
\caption{Full reconstruction efficiency under the common protocol.  ZipTok3D
uses five refinement passes.}
\label{tab:supp-reconstruction-efficiency}
\normalsize
\setlength{\tabcolsep}{4pt}
\begin{tabular}{lrrr}
\toprule
Method & Tok. & Full shape/s & Peak GiB \\
\midrule
VecSet & 512 & 4.20 & 32.63 \\
COD-VAE & 2 & 82.48 & 2.08 \\
COD-VAE & 32 & 79.82 & 2.08 \\
COD-VAE & 64 & 77.29 & 2.09 \\
\midrule
\ours{} & 1 & 72.51 & 2.03 \\
\ours{} & 2 & 72.51 & 2.03 \\
\ours{} & 4 & 72.24 & 2.03 \\
\bottomrule
\end{tabular}
\end{table}

Table~\ref{tab:supp-reconstruction-efficiency} reports the resulting
representation--compute tradeoff. Shorter prefixes reduce neither the dense
occupancy query nor the number of shared refinement calls. Consequently,
ZipTok3D has lower end-to-end throughput than the single-pass COD-VAE variants,
despite its shorter latent sequence. Peak memory remains similar because the
common encoder and dense-field query dominate the allocation.

\subsection{Statistical Analysis}

For the primary comparison between ZipTok3D at $K=4,L=5$ and COD-VAE-32, we
resample aligned TRELLIS objects with replacement. Each of 20,000 resamples
contains 2,613 object indices. For each metric, we compute the difference
between the paired sample means and report the 2.5th and 97.5th percentiles of
the resulting bootstrap distribution. The bootstrap uses seed 123456, and all
effects are defined as ZipTok3D minus COD-VAE, so negative CD is favorable.
These intervals quantify variation over evaluation objects conditional on the
trained models; they do not measure variation across independent training
runs.

\section{Post-Hoc Adaptive-Budget Diagnostic}
\label{sec:posthoc-oracle}

\begin{table}[!t]
\centering
\caption{All-metric post-hoc diagnostic. The \ours{} rows report the post-hoc
oracle. Found and Rate give the number and percentage of shapes for which the
search identifies an operating point that matches or exceeds COD-VAE-32 on all
three rounded per-shape metrics. Avg. $K$ and Avg. $L$ are computed over these
shapes. Metrics in the right block are aggregated over the same subset for
both methods, so the COD-VAE rows are subset rather than full-split results.
IoU and F1 are percentages.}
\label{tab:supp-posthoc-all}
{\normalsize\setlength{\tabcolsep}{2.5pt}
\begin{tabular}{@{}llccccccc@{}}
\toprule
& & \multicolumn{4}{c}{Post-hoc search} & \multicolumn{3}{c}{Subset metrics} \\
\cmidrule(lr){3-6}\cmidrule(l){7-9}
Data & Method & Found & Rate & Avg. $K$ & Avg. $L$
& IoU$\uparrow$ & CD$\downarrow$ & F1$\uparrow$ \\
\midrule
ShapeNet & COD-VAE-32 & -- & -- & 32.00 & -- & 97.2 & 0.011 & 98.1 \\
ShapeNet & \ours{} & 979/1,283 & 76.31 & 2.43 & 3.41 & 97.3 & 0.011 & 98.3 \\
\addlinespace[2pt]
TRELLIS & COD-VAE-32 & -- & -- & 32.00 & -- & 75.8 & 0.015 & 96.1 \\
TRELLIS & \ours{} & 1,767/2,613 & 67.62 & 3.02 & 3.97 & 76.6 & 0.014 & 96.6 \\
\bottomrule
\end{tabular}}
\end{table}

\begin{figure}[!t]
    \centering
    \captionsetup{skip=3pt}
    \begin{minipage}[t]{.485\textwidth}
        \centering
        \includegraphics[width=\linewidth]
        {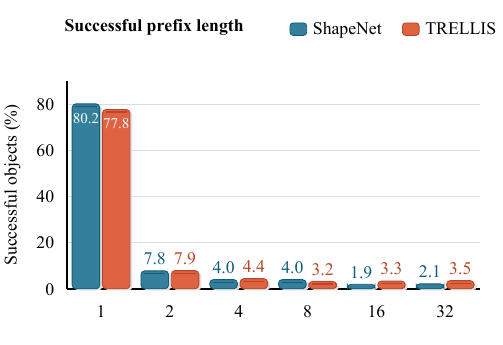}
        \par\smallskip
        \textbf{(a)} First successful prefix length $K$
    \end{minipage}
    \hfill
    \begin{minipage}[t]{.485\textwidth}
        \centering
        \includegraphics[width=\linewidth]
        {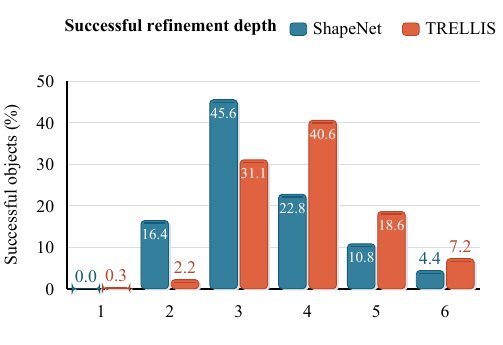}
        \par\smallskip
        \textbf{(b)} First successful refinement depth $L$
    \end{minipage}
    \caption{Adaptive-budget distributions under the token-first all-metric
    search. The left panel shows the first successful prefix length $K$ and the
    right panel the first successful refinement depth $L$; percentages are
    normalized within the successful subset of each dataset.}
    \label{fig:supp-adaptive-distributions}
\end{figure}

To characterize how the required budget varies across individual shapes, we
perform a post-hoc diagnostic over
$K\in\{1,2,4,8,16,32\}$ and $L\in\{1,\ldots,6\}$ from the same checkpoint.
Candidates are searched by increasing $K$ and then increasing $L$, prioritizing
a shorter representation before shallower decoding.  This diagnostic accesses
ground-truth reconstruction metrics and thus measures adaptive potential rather
than the performance of a deployable budget predictor.

We use a strict criterion that accepts a candidate only when
all three metrics are no worse than COD-VAE-32 after decimal half-up rounding
(one decimal for IoU/F1 and three decimals for CD).  Shapes without a successful
candidate are excluded from this diagnostic.  Table~\ref{tab:supp-posthoc-all}
reports both methods on the same successful subset, preventing differences in
subset composition from affecting the comparison.

No candidate in the evaluated grid satisfies the strict all-metric criterion
for the remaining 304 ShapeNet and 846 TRELLIS objects. All fixed-budget
evaluations include these objects; they are excluded only from the oracle
averages in Table~\ref{tab:supp-posthoc-all}. Consequently, the
reported average budgets characterize the successful subset rather than the
expected cost of a deployable early-exit policy.

Figure~\ref{fig:supp-adaptive-distributions} summarizes the first successful
budgets within this subset. Three or four passes account for 68.3\% and 71.6\% of the
ShapeNet and TRELLIS subsets, respectively. This distribution is consistent
with the dataset-level saturation behavior in
Figure~\ref{fig:supp-surface-curves}.

\makeatletter
\setlength{\@fptop}{0pt}
\makeatother

\begin{figure}[!t]
\centering
\includegraphics[width=.78\textwidth]
{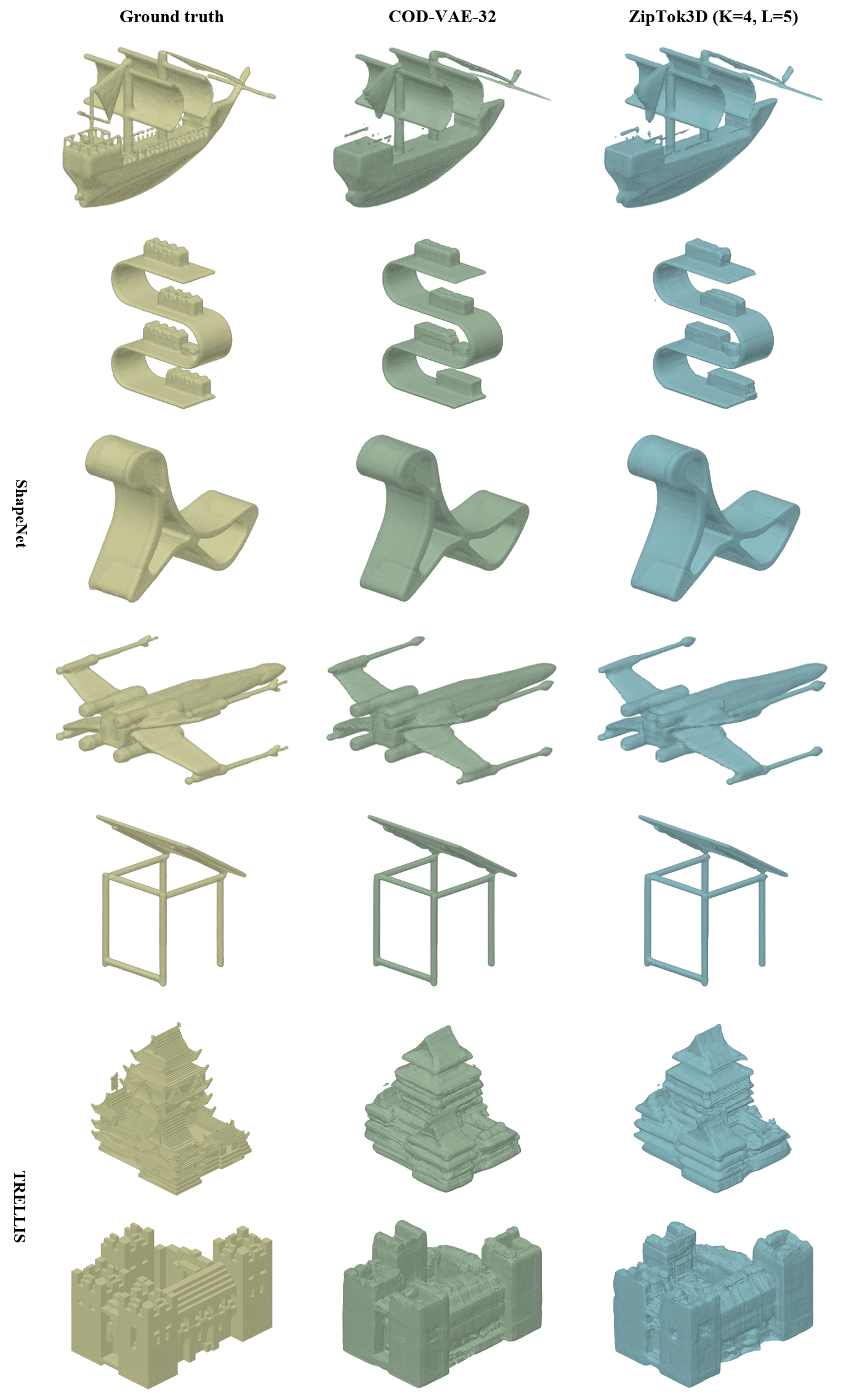}
\captionsetup{width=\textwidth,skip=3pt}
\caption{Additional reconstruction comparisons on ShapeNet (top five rows)
and TRELLIS (bottom two rows). ZipTok3D uses $K=4$ and $L=5$; the baseline
uses 32 COD-VAE tokens.}
\label{fig:supp-additional-qualitative}
\end{figure}

\section{Additional Qualitative Comparisons}

Figure~\ref{fig:supp-additional-qualitative} extends the qualitative comparison
to objects with thin components, large openings, curved surfaces, and repeated
architectural elements. The open shelf, chair, and table retain their principal
empty regions, while the vessel and aircraft preserve separated components and
overall topology. On TRELLIS, both reconstructions recover the multi-level
building layouts and repeated roof or fortification structures. The remaining
differences are concentrated around small protrusions, thin supports, roof
eaves, and crenellations, which remain challenging under both token budgets.

\clearpage

\makeatletter
\setlength{\@fptop}{0pt plus 1fil}
\makeatother

\end{document}